\documentclass{article} 
\usepackage{iclr2016_conference,times}
\usepackage{hyperref}
\usepackage{url}
\usepackage{amsmath,amssymb,float}
\usepackage{multirow}
\usepackage{tabularx}
\usepackage{todonotes}
\usepackage{caption}
\usepackage{subcaption}
\DeclareMathOperator{\E}{\mathbb{E}}
\def\!#1{\boldsymbol{#1}}
\def\*#1{\mathbf{#1}}

\title{The Variational Fair Autoencoder}
  
\author{Christos Louizos$^*$, Kevin Swersky$^\times$, Yujia Li$^\times$, Max Welling$^{*\dagger\ddagger}$, Richard Zemel$^{\times\dagger}$\\
	$^*$ Machine Learning Group, University of Amsterdam\\
	$^\times$Department of Computer Science, University of Toronto\\
	$^\dagger$ Canadian Institute for Advanced Research (CIFAR)\\
	$^\ddagger$ University of California, Irvine\\
	\texttt{C.Louizos@uva.nl, \{kswersky, yujiali\}@cs.toronto.edu}\\
	\texttt{M.Welling@uva.nl, zemel@cs.toronto.edu}}

\iclrfinalcopy 

\begin{document}

\maketitle

\begin{abstract}
We investigate the problem of learning representations that are invariant to certain nuisance or sensitive factors of variation in the data while retaining as much of the remaining information as possible. Our model is based on a variational autoencoding architecture~\citep{kingma2013auto, rezende2014stochastic} with priors that encourage independence between sensitive and latent factors of variation. Any subsequent processing, such as classification, can then be performed on this purged latent representation. To remove any remaining dependencies we incorporate an additional penalty term based on the ``Maximum Mean Discrepancy'' (MMD)~\citep{gretton2006kernel} measure. We discuss how these architectures can be efficiently trained on data and show in experiments that this method is more effective than previous work in removing unwanted sources of variation while maintaining informative latent representations. 
\end{abstract}

\section{Introduction} 
In ``Representation Learning'' one tries to find representations of the data that are informative for a particular task while removing the factors of variation that are uninformative and are typically detrimental for the task under consideration. Uninformative dimensions are often called ``noise'' or ``nuisance variables'' while informative dimensions are usually called latent or hidden factors of variation. Many machine learning algorithms can be understood in this way: principal component analysis, nonlinear dimensional reduction and latent Dirichlet allocation are all models that extract informative factors (dimensions, causes, topics) of the data which can often be used to visualize the data. On the other hand, linear discriminant analysis and deep (convolutional) neural nets learn representations that are good for classification. 

In this paper we  consider the case where we wish to learn latent representations where (almost) all of the information about certain known factors of variation are purged from the representation while still retaining as much information about the data as possible. In other words, we want a latent representation $\*z$ that is maximally informative about an observed random variable $\*y$ (e.g., class label) while minimally informative about a \emph{sensitive} or \emph{nuisance} variable $\*s$. By treating $\*s$ as a sensitive variable, i.e. $\*s$ is correlated with our objective, we are dealing with ``fair representations'', a problem previously considered by~\cite{zemel2013learning}. If we instead treat $\*s$ as a nuisance variable we are dealing with ``domain adaptation'', in other words by removing the domain $\*s$ from our representations we will obtain \emph{improved} performance.

In this paper we introduce a novel model based on deep variational autoencoders (VAE)~\citep{kingma2013auto, rezende2014stochastic}. These models can naturally encourage separation between latent variables $\*z$ and sensitive variables $\*s$ by using factorized priors $p(\*s)p(\*z)$. However, some dependencies may still remain when mapping data-cases to their hidden representation using the variational posterior $q(\*z|\*x,\*s)$, which we stamp out using a ``Maximum Mean Discrepancy"~\citep{gretton2006kernel} term that penalizes differences between all order moments of the marginal posterior distributions $q(\*z|\*s=k)$ and $q(\*z|\*s=k')$ (for a discrete RV $\*s$). In experiments we show that this combined approach is highly successful in learning representations that are devoid of unwanted information while retaining as much information as possible from what remains. 

\section{Learning Invariant Representations}
\begin{figure}[ht]
    \begin{center}
        \begin{minipage}[b]{0.48\textwidth}
            \centering
            \includegraphics[scale=0.75]{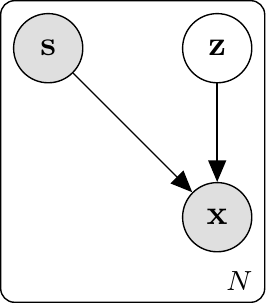}
            \caption{Unsupervised model}
        \end{minipage}
        \quad
          \begin{minipage}[b]{0.48\textwidth}
            \centering
            \includegraphics[scale=0.75]{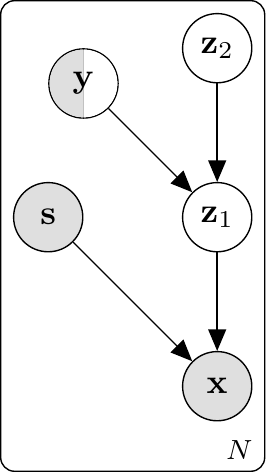}
            \caption{Semi-supervised model}
        \end{minipage}
    \end{center}
\end{figure}

\subsection{Unsupervised model}
Factoring out undesired variations from the data can be easily formulated as a general probabilistic model which admits two distinct (independent) ``sources''; an observed variable $\*s$, which denotes the variations that we want to remove, and a continuous latent variable $\*z$ which models all the remaining information.  This generative process can be formally defined as:
\begin{align*}
	\*z \sim p(\*z); \qquad \*x \sim p_\theta(\*x| \*z, \*s)
\end{align*}
where $p_\theta(\*x| \*z, \*s)$ is an appropriate probability distribution for the data we are modelling. With this formulation we explicitly encode a notion of `invariance' in our model, since the latent representation is marginally independent of the factors of variation $\*s$. Therefore the problem of finding an invariant representation for a data point $\*x$ and variation $\*s$ can be cast as performing inference on this graphical model and obtaining the posterior distribution of $\*z$, $p(\*z|\*x, \*s)$. 

For our model we will employ a variational autoencoder architecture~\citep{kingma2013auto,rezende2014stochastic}; namely we will parametrize the generative model (decoder) $p_\theta(\*x|\*z,\*s)$ and  the variational posterior (encoder) $q_\phi(\*z|\*x,\*s)$ as (deep) neural networks which accept as inputs $\*z,\*s$ and $\*x,\*s$ respectively and produce the parameters of each distribution after a series of non-linear transformations. Both the model ($\theta$) and variational ($\phi$) parameters will be jointly optimized with the SGVB~\citep{kingma2013auto} algorithm according to a lower bound on the log-likelihood. This parametrization will allow us to capture most of the salient information of $\*x$ in our embedding $\*z$. Furthermore the distributed representation of a neural network would allow us to better resolve the dependencies between $\*x$ and $\*s$ thus yielding a better disentangling between the independent factors $\*z$ and $\*s$. By choosing a Gaussian posterior $q_\phi(\*z|\*x, \*s)$ and standard isotropic Gaussian prior $p(\*z) = \mathcal{N}(\*0, \*I)$ we can obtain the following lower bound:
\begin{align}
	\sum_{n=1}^{N}\log p(\*x_n|\*s_n) & \ge  \sum_{n=1}^{N}\E_{q_\phi(\*z_n|\*x_n,\*s_n)}[\log p_\theta(\*x_n|\*z_n,\*s_n)] - KL(q_\phi(\*z_n|\*x_n,\*s_n) || p(\*z))\\
	                    & = \mathcal{F}(\phi, \theta; \*x_n, \*s_n) \nonumber
\end{align}
with $q_\phi(\*z_n|\*x_n, \*s_n) = \mathcal{N}(\*z_n|\!\mu_n = f_\phi(\*x_n, \*s_n), \!\sigma_n = e^{f_\phi(\*x_n, \*s_n)})$ and $p_\theta(\*x_n|\*z_n, \*s_n) = f_\theta(\*z_n, \*s_n)$ with $f_\theta(\*z_n, \*s_n)$ being an appropriate probability distribution for the data we are modelling.

\subsection{Semi-Supervised model}
Factoring out variations in an unsupervised way can however be harmful in cases where we want to use this invariant representation for a subsequent prediction task. In particular if we have a situation where the nuisance variable $\*s$ and the actual label $\*y$ are correlated, then training an unsupervised model could yield \emph{random} or \emph{degenerate} representations with respect to $\*y$. Therefore it is more appropriate to try to ``inject'' the information about the label during the feature extraction phase. This can be quite simply achieved by introducing a second ``layer'' of latent variables to our generative model where we try to correlate $\*z$ with the prediction task. Assuming that the invariant features are now called $\*z_1$ we enrich the generative story by similarly providing two distinct (independent) sources for $\*z_1$; a discrete (in case of classification)variable $\*y$ which denotes the label of the data point $\*x$ and a continuous latent variable $\*z_2$ which encodes the variation on $\*z_1$ that is not explained by $\*y$ ($\*x$ dependent noise). The process now can be formally defined as: 
\begin{align*}
\*y, \*z_2 \sim \text{Cat}(\*y)p(\*z_2); \qquad  \*z_1 \sim p_\theta(\*z_1|\*z_2, \*y);\qquad \*x \sim p_\theta(\*x| \*z_1,\*s)\nonumber
\end{align*}   
Similarly to the unsupervised case we use a variational auto-encoder and jointly optimize the variational and model parameters. The lower bound now becomes:
\begin{align}
\sum_{n=1}^{N}\log p(\*x_n|\*s_n) & \ge \sum_{n=1}^{N}\E_{q_\phi({\*z_1}_n, {\*z_2}_n, \*y_n| \*x_n, \*s_n)}[\log p(\*z_2) + \log p(\*y_n) + \log p_\theta({\*z_1}_n|{\*z_2}_n, \*y_n) + \nonumber\\&\qquad\qquad\qquad\qquad + \log p_\theta(\*x_n|{\*z_1}_n,\*s_n) - \log q_\phi({\*z_1}_n, {\*z_2}_n, \*y_n|\*x_n, \*s_n)]  
\end{align}
where we assume that the posterior $q_\phi({\*z_1}_n, {\*z_2}_n, \*y_n|\*x_n, \*s_n)$ is factorized as $q_\phi({\*z_1}_n, {\*z_2}_n, \*y_n|\*x_n, \*s_n) = q_\phi({\*z_1}_n|\*x_n, \*s_n)q_\phi(\*y_n|{\*z_1}_n)q_\phi({\*z_2}_n|{\*z_1}_n,\*y_n)$, and where:
\begin{align*}
q_\phi({\*z_1}_n|\*x_n, \*s_n) & = \mathcal{N}({\*z_1}_n|\!\mu_n = f_\phi(\*x_n, \*s_n), \!\sigma_n = e^{f_\phi(\*x_n, \*s_n)}) \\
q_\phi(\*y_n | {\*z_1}_n) & = \text{Cat}(\*y_n|\!\pi_n = \text{softmax}(f_\phi({\*z_1}_n)))\\
q_\phi({\*z_2}_n|{\*z_1}_n, \*y_n) & = \mathcal{N}({\*z_2}_n| \!\mu_n = f_\phi({\*z_1}_n, \*y_n), \!\sigma_n = e^{f_\phi({\*z_1}_n, \*y_n)})\\
p_\theta({\*z_1}_n | {\*z_2}_n, \*y_n) & = \mathcal{N}({\*z_1}_n| \!\mu_n = f_\theta({\*z_2}_n, \*y_n), \!\sigma_n = e^{f_\theta({\*z_2}_n, \*y_n)})\\
p_\theta(\*x_n|{\*z_1}_n, \*s_n) & = f_\theta({\*z_1}_n, \*s_n)
\end{align*}
with $f_\theta({\*z_1}_n, \*s_n)$ again being an appropriate probability distribution for the data we are modelling. The model proposed here can be seen as an extension to the `stacked M1+M2' model originally proposed from~\cite{kingma2014semi}, where we have additionally introduced the nuisance variable $\*s$ during the feature extraction. Thus following~\cite{kingma2014semi} we can also handle the `semi-supervised' case, i.e., missing labels. In situations where the label is observed the lower bound takes the following form (exploiting the fact that we can compute some Kullback-Leibler divergences explicitly in our case):
\begin{align}
\sum_{n=1}^{N}\mathcal{L}_s(\phi, \theta; \*x_n, \*s_n, \*y_n) & =\sum_{n=1}^{N_s} \E_{q_\phi({\*z_1}_n|\*x_n, \*s_n)}[-KL(q_\phi({\*z_2}_n|{\*z_1}_n, \*y_n) || p(\*z_2)) + \log p_\theta(\*x_n|{\*z_1}_n, \*s_n)] +\nonumber\\&\qquad\qquad + \E_{q_\phi({\*z_1}_n|\*x_n, \*s_n)q_\phi({\*z_2}_n|{\*z_1}_n,\*y_n)}[\log p_\theta({\*z_1}|{\*z_2}_n, \*y_n)- \log q_\phi({\*z_1}_n|\*x_n \*s_n)]
\end{align}
and in the case that it is not observed we use $q(\*y_n|{\*z_1}_n)$ to `impute' our data:
\begin{align}
\sum_{m=1}^{M}\mathcal{L}_u(\phi, \theta; \*x_m, \*s_m) & = \sum_{m=1}^{M}\E_{q_\phi({\*z_1}_m|\*x_m,\*s_m)}[ - KL (q(\*y_m|{\*z_1}_m) || p(\*y_m)) + \log p_\theta(\*x_m|{\*z_1}_m, \*s_m)] +\nonumber\\&\qquad\qquad +  \E_{q_\phi({\*z_1}_m, \*y_m|\*x_m,\*s_m)}[ - KL (q_\phi({\*z_2}_m|{\*z_1}_m,\*y_m) || p(\*z_2))] +\nonumber\\&\qquad\qquad+  \E_{q_\phi({\*z_1}_m, \*y_m, {\*z_2}_m|\*x_m, \*s_m)}[\log p_\theta({\*z_1}_m|{\*z_2}_m,\*y_m) -\log q_\phi({\*z_1}_m|\*x_m,\*s_m) ]
\end{align}
therefore the final objective function is:
\begin{align}
    \mathcal{F}_{\text{VAE}}(\phi, \theta; \*x_n, \*x_m, \*s_n, \*s_m, \*y_n) & = \sum_{n=1}^{N}\mathcal{L}_s(\phi, \theta; \*x_n, \*s_n, \*y_n) + \sum_{m=1}^{M}\mathcal{L}_u(\phi, \theta; \*x_m, \*s_m) + \nonumber \\ &\qquad\qquad +  \alpha\sum_{n=1}^{N}\E_{q({\*z_1}_n|\*x_n, \*s_n)}[- \log q_\phi(\*y_n|{\*z_1}_n)]
\end{align}
where the last term is introduced so as to ensure that the predictive posterior $q_\phi(\*y|\*z_1)$ learns from both labeled and unlabeled data. This semi-supervised model will be called ``VAE'' in our experiments.

However, there is a subtle difference between the approach of~\cite{kingma2014semi} and our model. Instead of training separately each layer of stochastic variables we optimize the model jointly. The potential advantages of this approach are two fold: as we previously mentioned if the label $\*y$ and the nuisance information $\*s$ are correlated then training a (conditional) feature extractor separately poses the danger of creating a degenerate representation with respect to the label $\*y$. Furthermore the label information will also better guide the feature extraction towards the more salient parts of the data, thus maintaining most of the (predictive) information.

\subsection{Further invariance via Maximum Mean Discrepancy}
Despite the fact that we have a model that encourages statistical independence between $\*s$ and $\*z_1$ a-priori we might still have some dependence in the (approximate) marginal posterior $q_{\phi}(\*z_1|\*s)$. In particular, this can happen if the label $\*y$ is correlated with the sensitive variable $\*s$, which can allow information about $\*s$ to ``leak'' into the posterior. Thus instead we could maximize a ``penalized'' lower bound where we impose some sort of regularization on the marginal $q_{\phi}(\*z_1|\*s)$. In the following we will describe one way to achieve this regularization through the Maximum Mean Discrepancy (MMD)~\citep{gretton2006kernel} measure.

\subsubsection{Maximum Mean Discrepancy}
Consider the problem of determining whether two datasets $\{ \*X \} \sim P_0$ and $\{ \*X' \} \sim P_1$ are drawn from the same distribution, i.e., $P_0 = P_1$. A simple test is to consider the distance between empirical statistics $\psi(\cdot)$ of the two datasets:
\begin{equation}
    \left \| \frac{1}{N_0} \sum_{i=1}^{N_0} \psi(\*x_i) - \frac{1}{N_1} \sum_{i=1}^{N_1} \psi(\*x'_i) \right \|^2. \label{eq:premmd}
\end{equation}
Expanding the square yields an estimator composed only of inner products on which the kernel trick can be applied. The resulting estimator is known as Maximum Mean Discrepancy (MMD)~\citep{gretton2006kernel}:
\begin{align}
    \ell_{\mathrm{MMD}}(\*X, \*X') &= \frac{1}{N_0^2} \sum_{n=1}^{N_0} \sum_{m=1}^{N_0} k(\*x_n, \*x_{m}) + \frac{1}{N_1^2} \sum_{n=1}^{N_1} \sum_{m=1}^{N_1} k(\*x'_n, \*x'_m) - \frac{2}{N_0 N_1} \sum_{n=1}^{N_0} \sum_{m=1}^{N_1} k(\*x_n, \*x'_m). \label{eq:mmd}
\end{align}
Asymptotically, for a universal kernel such as the Gaussian kernel $k(x,x')=e^{-\gamma \| \*x - \*x' \|^2}$, $\ell_{\mathrm{MMD}}(\*X, \*X')$ is $0$ if and only if $P_0 = P_1$. Equivalently, minimizing MMD can be viewed as matching all of the moments of $P_0$ and $P_1$.  Therefore, we can use it as an extra ``regularizer'' and force the model to try to match the moments between the marginal posterior distributions of our latent variables, i.e., $q_{\phi}(\*z_1|s=0)$ and $q_{\phi}(\*z_1| s=1)$ (in the case of binary nuisance information $\*s$\footnote{In case that we have more than two states for the nuisance information $\*s$, we minimize the MMD penalty between each marginal $q(\*z|\*s=k)$ and $q(\*z)$, i.e., $\sum_{k=1}^{K}\ell_{\mathrm{MMD}}({\*Z_1}_{\*s=k}, {\*Z_1})$ for all possible states $K$ of $\*s$.}). By adding the MMD penalty into the lower bound of our aforementioned VAE architecture we obtain our proposed model, the ``Variational Fair Autoencoder'' (VFAE):
\begin{align}
    \mathcal{F}_{\text{VFAE}}(\phi, \theta; \*x_n, \*x_m, \*s_n, \*s_m, \*y_n) & = \mathcal{F}_{\text{VAE}}(\phi, \theta; \*x_n, \*x_m, \*s_n, \*s_m, \*y_n) -\beta \ell_{\mathrm{MMD}}({\*Z_1}_{\*s=0}, {\*Z_1}_{\*s=1})
\end{align}
where:
\begin{align}
  \ell_{\mathrm{MMD}}({\*Z_1}_{\*s=0}, {\*Z_1}_{\*s=1}) & = \|\E_{\tilde{p}(\*x|\*s=0)}[\E_{q(\*z_1|\*x, \*s=0)}[\psi(\*z_1)]] - E_{\tilde{p}(\*x|\*s=1)}[\E_{q(\*z_1|\*x, \*s=1)}[\psi(\*z_1)]]\|^2
\end{align}

\subsection{Fast MMD via Random Fourier Features}
A naive implementation of MMD in minibatch stochastic gradient descent would require computing the $M\times M$ Gram matrix for each minibatch during training, where $M$ is the minibatch size. Instead, we can use random kitchen sinks~\citep{rahimi2009weighted} to compute a feature expansion such that computing the estimator $(\ref{eq:premmd})$ approximates the full MMD (\ref{eq:mmd}). To compute this, we draw a random $K\times D$ matrix $\*W$, where $K$ is the dimensionality of $\*x$, $D$ is the number of random features and each entry of $\*W$ is drawn from a standard isotropic Gaussian.
The feature expansion is then given as:
\begin{align}
\psi_\*W(\*x) &= \sqrt{\frac{2}{D}}\mathrm{cos}\left ( \sqrt{\frac{2}{\gamma}} \*x\*W + \*b \right ).
\end{align}
where $\*b$ is a $D$-dimensional uniform random vector with entries in $[0,2\pi]$. \cite{zhao2014fastmmd} have successfully applied the idea of using random kitchen sinks to approximate MMD. This estimator is fairly accurate, and is typically much faster than the full MMD penalty. We use $D=500$ in our experiments.

\section{Experiments}
We performed experiments on the three datasets that correspond to a ``fair'' classification scenario and were previously used by~\cite{zemel2013learning}. In these datasets the ``nuisance'' or sensitive variable $\*s$ is significantly correlated with the label $\*y$ thus making the proper removal of $\*s$ challenging. Furthermore, we also experimented with the Amazon reviews dataset to make a connection with the ``domain-adaptation'' literature. Finally, we also experimented with a more general task on the extended Yale B dataset; that of learning invariant representations. 

\subsection{Datasets}
For the fairness task we experimented with three datasets that were previously used by~\citet{zemel2013learning}. The German dataset is the smallest one with $1000$ data points and the objective is to predict whether a person has a good or bad credit rating. The sensitive variable is the gender of the individual. The Adult income dataset contains $45,222$ entries and describes whether an account holder has over $\$50,000$ dollars in their account. The sensitive variable is age. Both of these are obtained from the UCI machine learning repository~\citep{UCI}. The health dataset is derived from the Heritage Health Prize\footnote{\url{www.heritagehealthprize.com}}. It is the largest of the three datasets with $147,473$ entries. The task is to predict whether a patient will spend any days in the hospital in the next year and the sensitive variable is the age of the individual. We use the same train/test/validation splits as~\citet{zemel2013learning} for our experiments. Finally we also binarized the data and used a multivariate Bernoulli distribution for $p_\theta(\*x_n|{\*z_1}_n,\*s_n) = \text{Bern}(\*x_n | \!\pi_n = \sigma(f_\theta({\*z_1}_n, \*s_n)))$, where $\sigma(\cdot)$ is the sigmoid function \footnote{$\sigma(t) = \frac{1}{1 + e^{-t}}$}.

For the domain adaptation task we used the Amazon reviews dataset (with similar preprocessing) that was also employed by~\citet{chen2012marginalized} and~\citet{2015arXiv150507818G}. It is composed from text reviews about particular products, where each product belongs to one out of four different domains: ``books'', ``dvd'', ``electronics'' and ``kitchen''. As a result we performed twelve domain adaptation tasks. The labels $\*y$ correspond to the sentiment of each review, i.e. either positive or negative. Since each feature vector $\*x$ is composed from counts of unigrams and bigrams we used a Poisson distribution for $p_\theta(\*x_n|{\*z_1}_n, \*s_n) = \text{Poisson}(\*x_n | \!\lambda_n = e^{f_\theta({\*z_1}_n, \*s_n)})$. It is also worthwhile to mention that we can fully exploit the semi-supervised nature of our model in this dataset, and thus for training we only use the source domain labels and consider the labels of the target domain as ``missing''.

For the general task of learning invariant representations we used the Extended Yale B dataset, which was also employed in a similar fashion by~\citet{li2014learning}. It is composed from face images of 38 people under different lighting conditions (directions of the light source). Similarly to~\citet{li2014learning}, we created 5 states for the nuisance variable $\*s$: light source in upper right, lower right, lower left, upper left and the front. The labels $\*y$ correspond to the identity of the person. Following~\citet{li2014learning}, we used the same training, test set and no validation set. For the $p(\*x_n|{\*z_1}_n, \*s_n)$ distribution we used a Gaussian with means constrained in the 0-1 range (since we have intensity images) by a sigmoid, i.e. $p_\theta(\*x_n|{\*z_1}_n, \*s_n) = \mathcal{N}(\*x_n| \!\mu_n = \sigma(f_\theta({\*z_1}_n, \*s_n)), \!\sigma_n = e^{f_\theta({\*z_1}_n, \*s_n)})$.

\subsection{Experimental Setup}
For the Adult dataset both encoders, for $\*z_1$ and $\*z_2$, and both decoders, for $\*z_1$ and $\*x$, had one hidden layer of 100 units. For the Health dataset we had one hidden layer of 300 units for the $\*z_1$ encoder and $\*x$ decoder and one hidden layer of 150 units for the $\*z_2$ encoder and $\*z_1$ decoder. For the much smaller German dataset we used 60 hidden units for both encoders and decoders. Finally, for the Amazon reviews and Extended Yale B datasets we had one hidden layer with 500, 400 units respectively for the $\*z_1$ encoder, $\*x$ decoder, and 300, 100 units respectively for the $\*z_2$ encoder and $\*z_1$ decoder. On all of the datasets we used 50 latent dimensions for $\*z_1$ and $\*z_2$, except for the small German dataset, where we used 30 latent dimensions for both variables. For the predictive posterior $q_\phi(\*y|\*z_1)$ we used a simple Logistic regression classifier. Optimization of the objective function was done with Adam~\citep{DBLP:journals/corr/KingmaB14} using the default values for the hyperparameters, minibatches of 100 data points and temporal averaging. The MMD penalty was simply multiplied by the minibatch size so as to keep the scale of the penalty similar to the lower bound. Furthermore, the extra strength of the MMD, $\beta$, was tuned according to a validation set. The scaling of the supervised cost was low ($\alpha = 1 $) for the Adult, Health and German datasets due to the correlation of $\*s$ with $\*y$. On the Amazon reviews and Extended Yale B datasets however the scaling of the supervised cost was higher: $\alpha = 100 \cdot \frac{N_{\text{batch\_source}} + N_{\text{batch\_target}}}{N_{\text{batch\_source}}}$ for the Amazon reviews dataset (empirically determined after observing the classification loss on the first few iterations on the first source-target pair) and $\alpha = 200$ for the Extended Yale B dataset. Similarly, the scaling of the MMD penalty was $\beta = 100 \cdot N_{\text{batch}}$ for the Amazon reviews dataset and $\beta = 200 \cdot N_{\text{batch}}$ for the Extended Yale B.

Our evaluation is geared towards two fronts; removing information about $\*s$ and classification accuracy for $\*y$. To measure the information about $\*s$ in our new representation we simply train a classifier to predict $\*s$ from $\*z_1$. We utilize both Logistic Regression (LR) which is a simple linear classifier, and Random Forest (RF) which is a powerful non-linear classifier. Since on the datasets that we experimented with the nuisance variable $\*s$ is binary we can easily find the random chance accuracy for $\*s$ and measure the discriminatory information of $\*s$ in $\*z_1$. Furthermore, we also used the discrimination metric from~\cite{zemel2013learning} as well a  more ``informed'' version of the discrimination metric that instead of the predictions, takes into account the probabilities of the correct class. They are provided in the appendix A. Finally, for the classification performance on $\*y$ we used the predictive posterior $q_\phi(\*y|\*z_1)$ for the VAE/VFAE and a simple Logistic Regression for the original representations $\*x$. It should be noted that for the VFAE and VAE models we use a sample from $q_\phi(\*z_1|\*x, \*s)$ to make predictions, instead of using the mean. We found that the extra noise helps with invariance. 

We implemented the Learning Fair Representations~\citep{zemel2013learning} method (LFR) as a baseline using $K=50$ dimensions for the latent space. To measure the accuracy on $\*y$ in the results below we similarly used the LFR model predictions.

\subsection{Results}
\subsubsection{Fair classification}
\begin{figure}[ht]
   \centering
   \begin{subfigure}{\linewidth}
  \begin{subfigure}{.329\textwidth}
    \centering
        \includegraphics[width=1.1\linewidth]{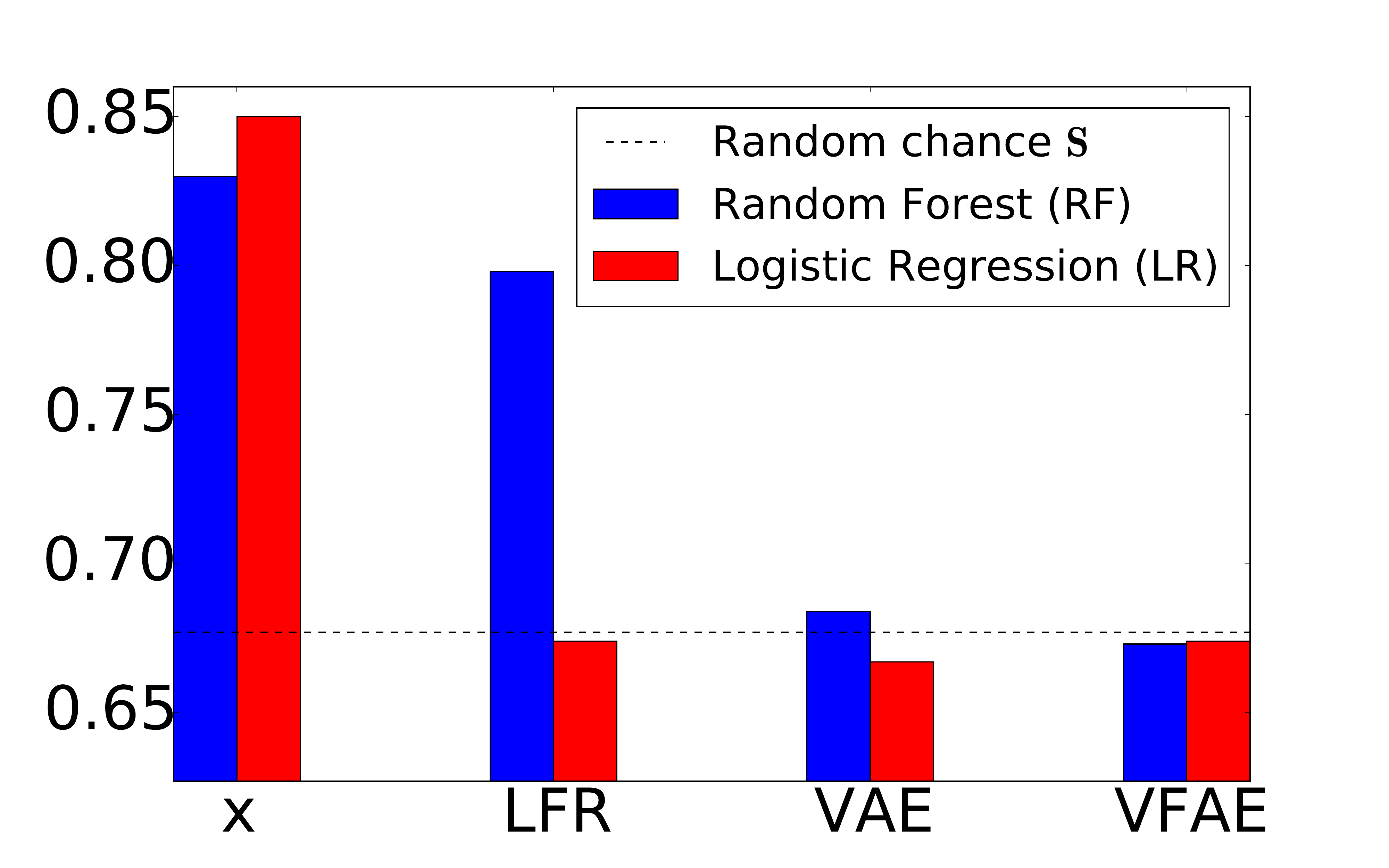}
  \end{subfigure} %
  \begin{subfigure}{.329\textwidth}
    \centering
        \includegraphics[width=1.1\linewidth]{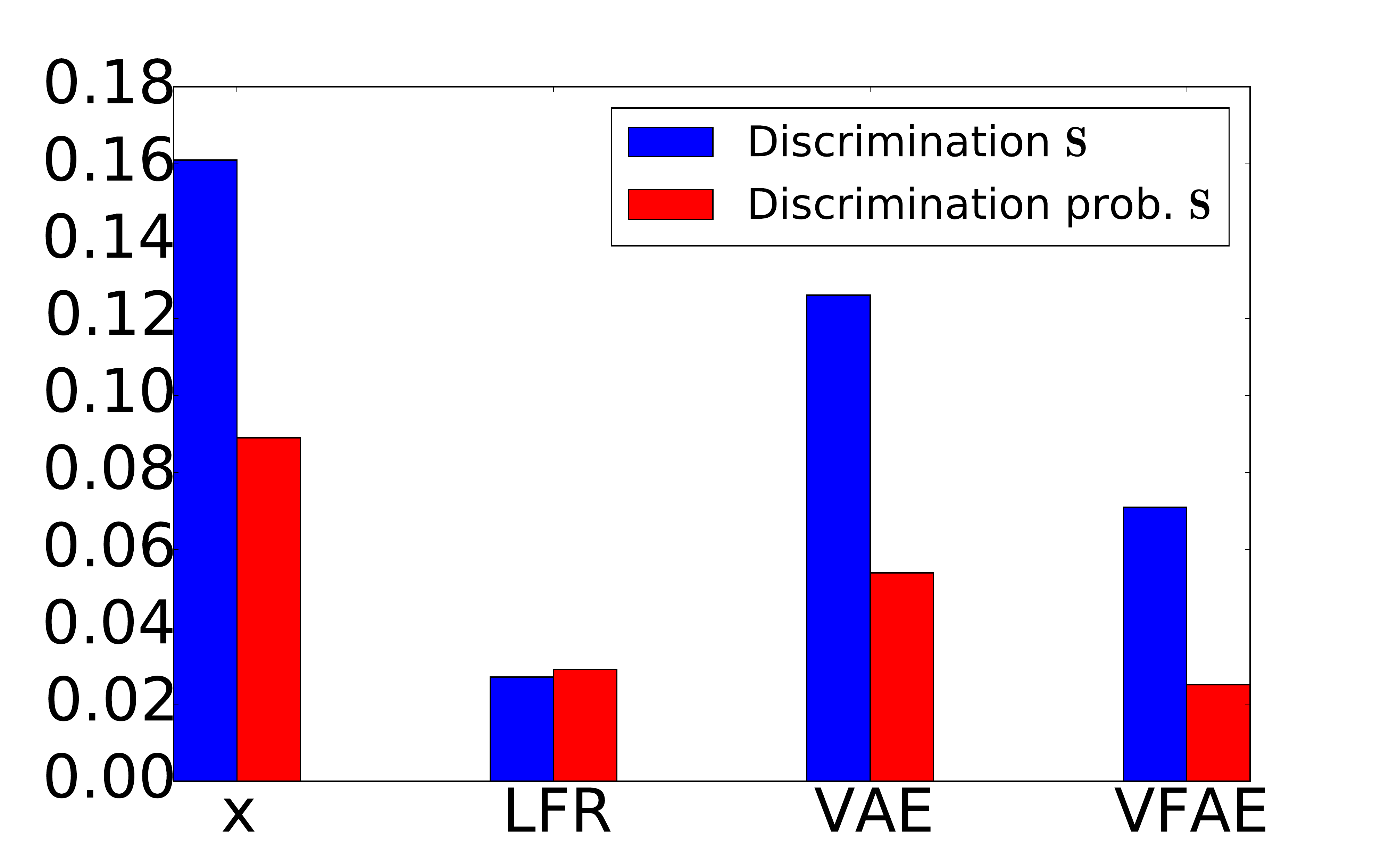}
  \end{subfigure} %
  \begin{subfigure}{.329\textwidth}
    \centering
        \includegraphics[width=1.1\linewidth]{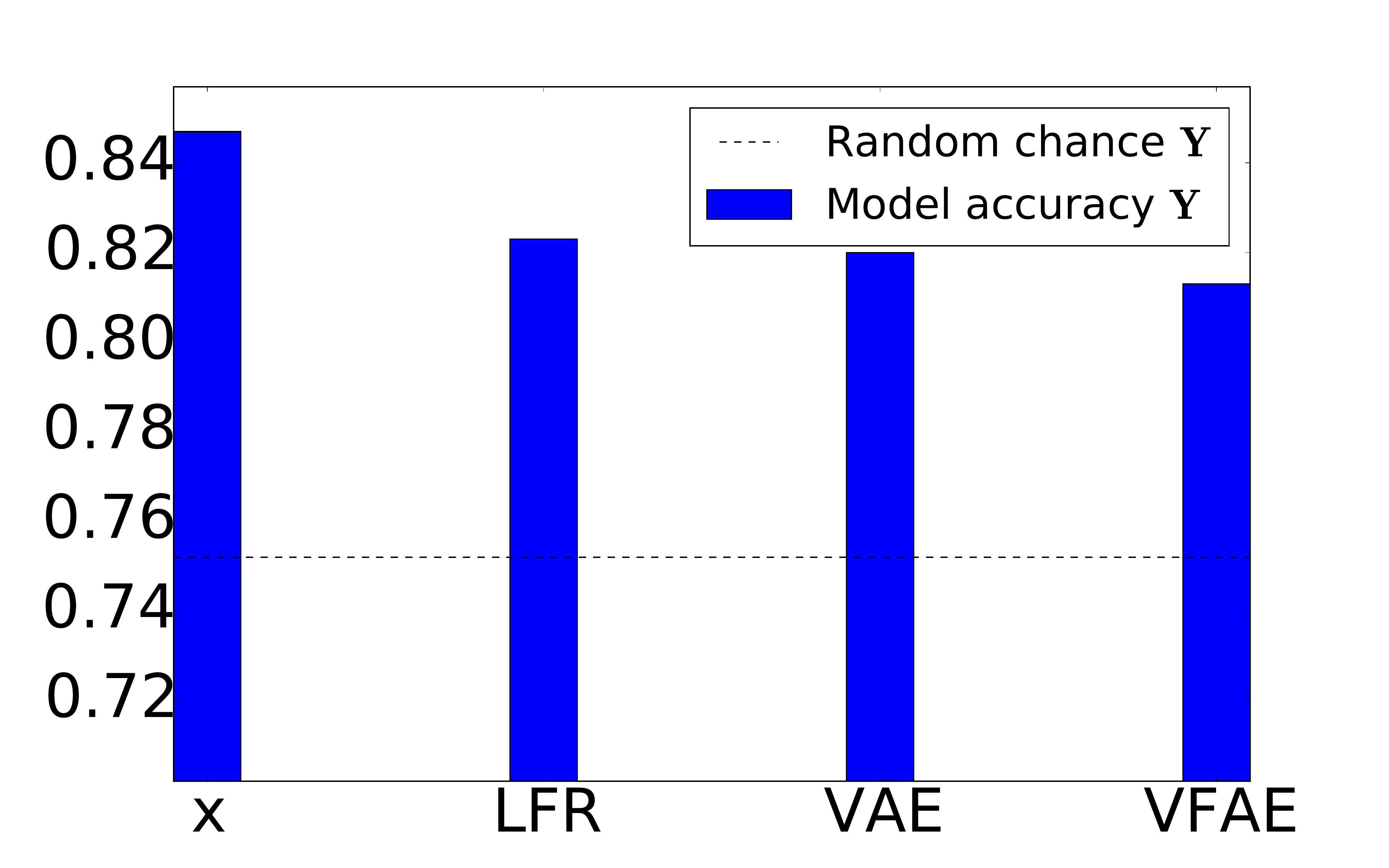}
  \end{subfigure}
  \caption{Adult dataset}
  \end{subfigure} \\
  \begin{subfigure}{\linewidth}
  \begin{subfigure}{.329\textwidth}
    \centering
        \includegraphics[width=1.1\linewidth]{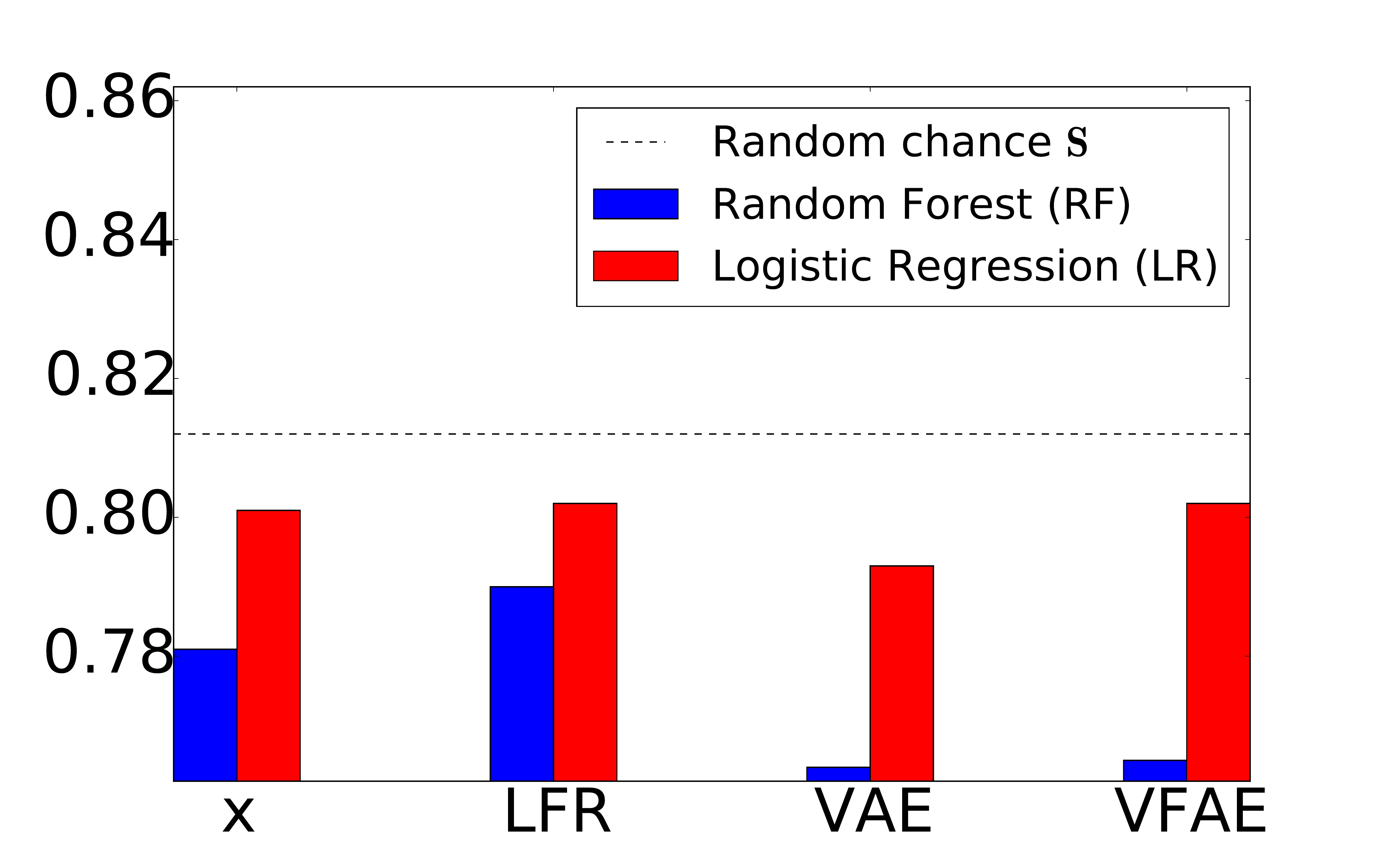}
  \end{subfigure}%
    \begin{subfigure}{.329\textwidth}
    \centering
        \includegraphics[width=1.1\linewidth]{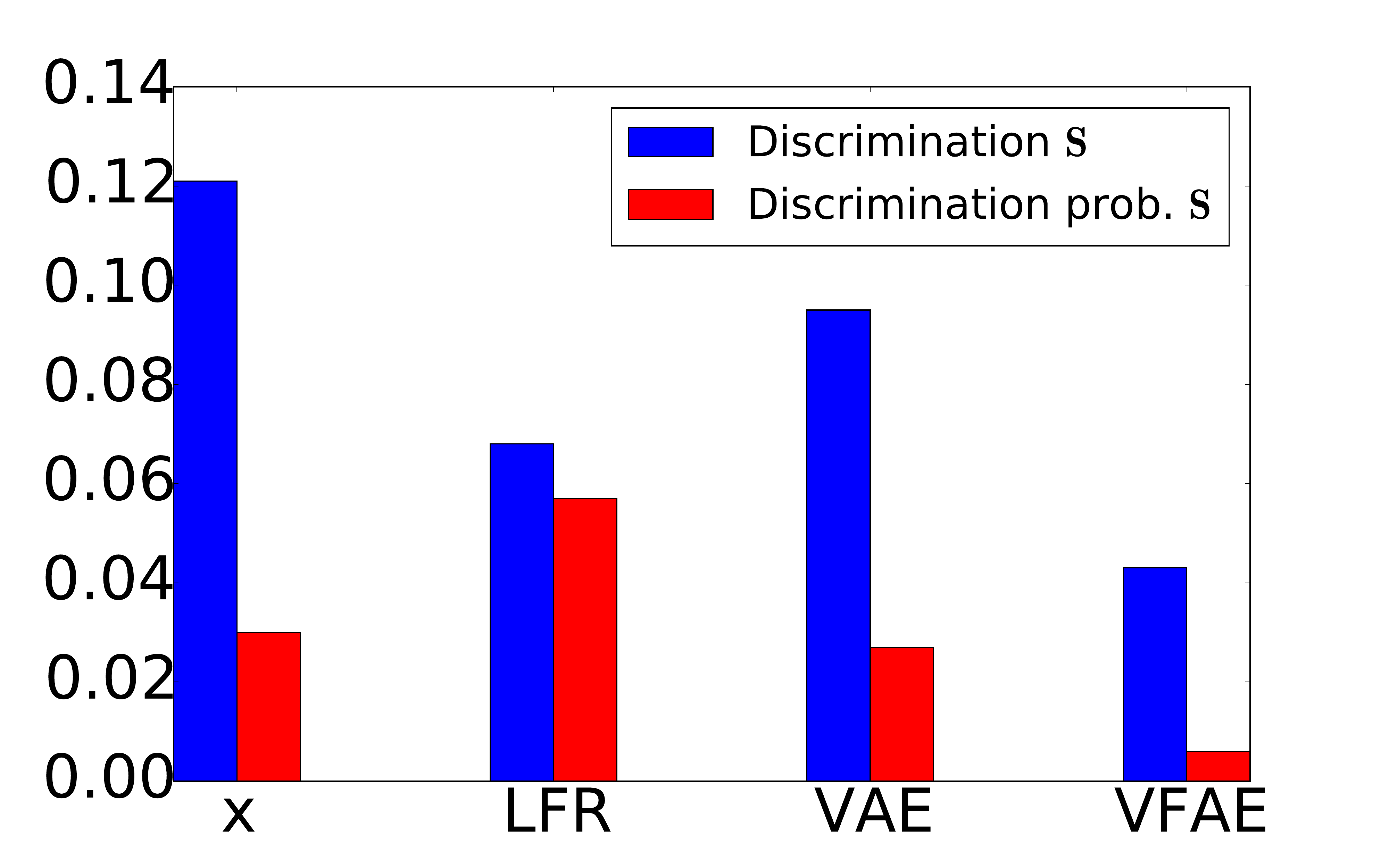}
  \end{subfigure}%
  \begin{subfigure}{.329\textwidth}
    \centering
        \includegraphics[width=1.1\linewidth]{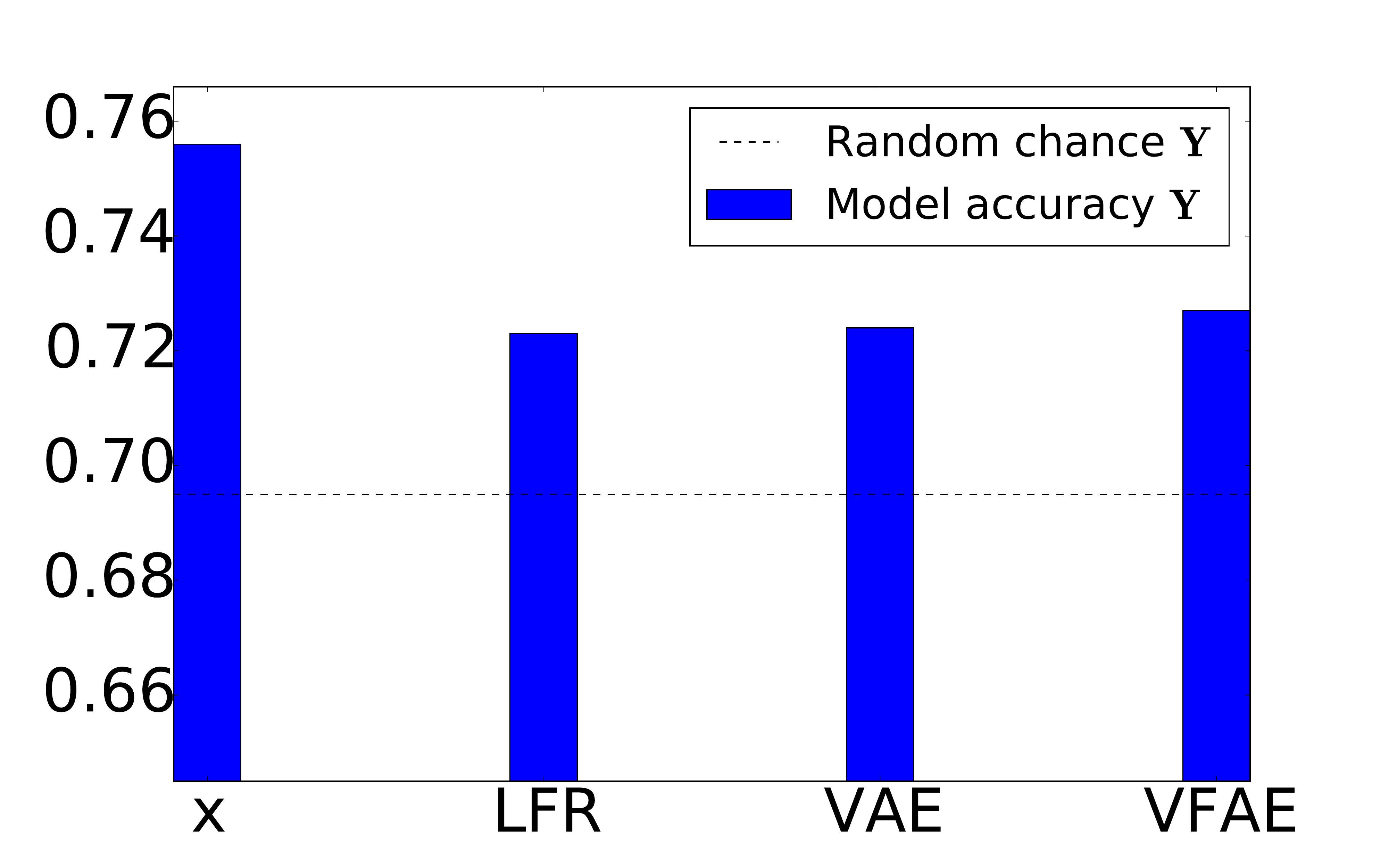}
  \end{subfigure}
  \caption{German dataset}
  \end{subfigure}\\
  \begin{subfigure}{\linewidth}
  \begin{subfigure}{.329\textwidth}
      \centering
      \includegraphics[width=1.1\linewidth]{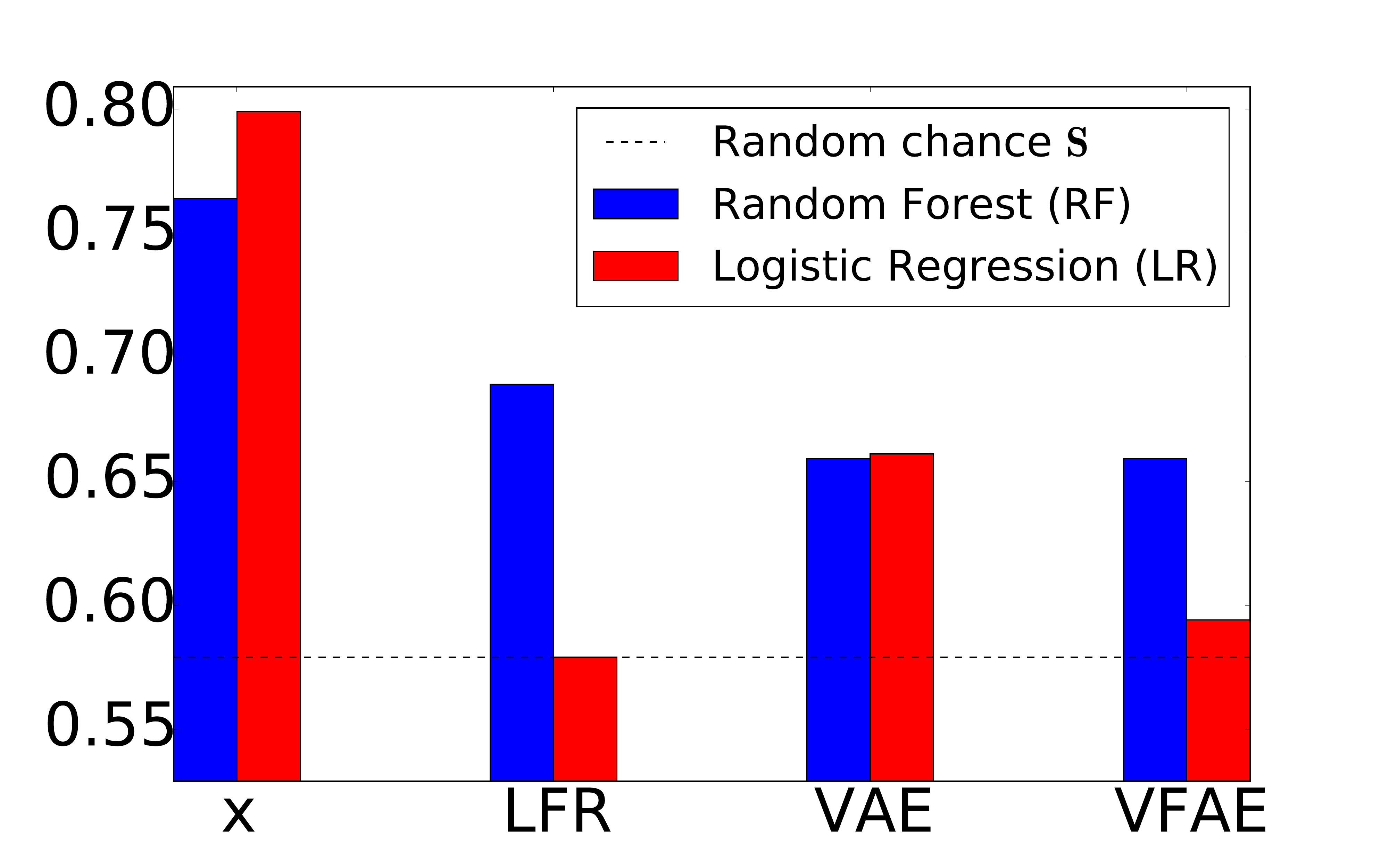}
  \end{subfigure}%
  \begin{subfigure}{.329\textwidth}
      \centering
      \includegraphics[width=1.1\linewidth]{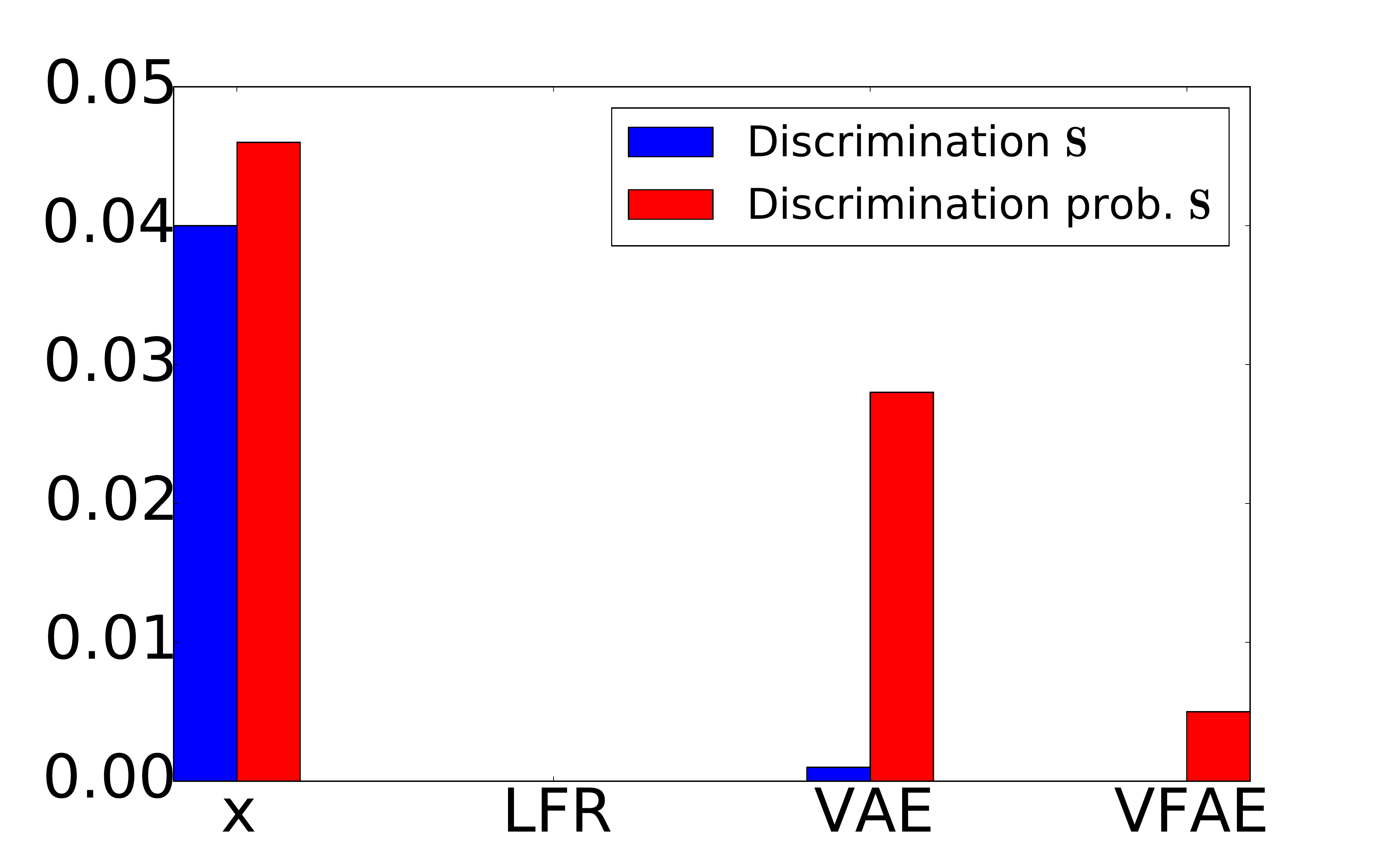}
  \end{subfigure}%
  \begin{subfigure}{.329\textwidth}
      \centering
      \includegraphics[width=1.1\linewidth]{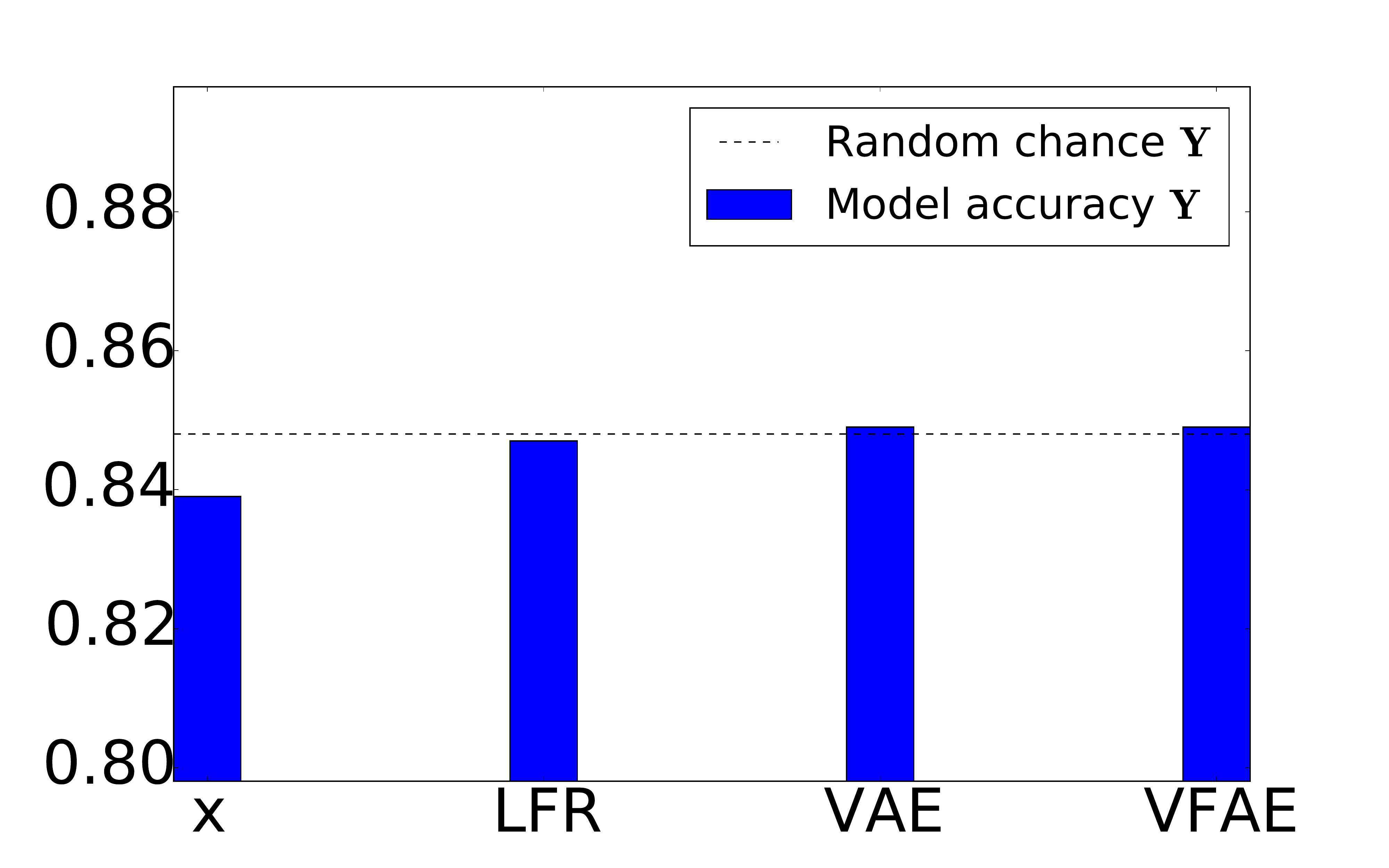}
  \end{subfigure}
  \caption{Health dataset}
  \end{subfigure}%
  \caption{Fair classification results. Columns correspond to each evaluation scenario (in order): Random/RF/LR accuracy on $\*s$, Discrimination/Discrimination prob. against $\*s$ and Random/Model accuracy on $\*y$. Note that the objective of a ``fair'' encoding is to have low accuracy on S (where LR is a linear classifier and RF is nonlinear), low discrimination against S and high accuracy on Y.}
  \label{tab:fair_results}
\end{figure}

The results for all three datasets can be seen in Figure~\ref{tab:fair_results}. Since we are dealing with the ``fair'' classification scenario here, low accuracy and discrimination against $\*s$ is more important than the accuracy on $\*y$ (as long as we do not produce degenerate representations).

On the Adult dataset, the highest accuracy on the label $\*y$ and the lowest discrimination against $\*s$ is obtained by our LFR baseline. Despite the fact that LFR appears to give the best tradeoff between accuracy and discrimination, it appears to retain information about $\*s$ in its representation, which is discovered from the random forest classifier. In that sense, the VFAE method appears to do the best job in actually removing the sensitive information and maintaining most of the predictive information. Furthermore, the introduction of the MMD penalty in the VFAE model seems to provide a significant benefit with respect to our discrimination metrics, as both were reduced considerably compared to the regular VAE. 

On the German dataset, all methods appear to be invariant with respect to the sensitive information $\*s$. However this is not the case for the discrimination metric, since LFR does appear to retain information compared to the VAE and VFAE. The MMD penalty in VFAE did seem improve the discrimination scores over the original VAE, while the accuracy on the labels $\*y$ remained similar. 

As for the Health dataset; this dataset is extremely imbalanced, with only 15\% of the patients being admitted to a hospital. Therefore, each of the classifiers seems to predict the majority class as the label $\*y$ for every point. For the invariance against $\*s$ however, the results were more interesting. On the one hand, the VAE model on this dataset did maintain some sensitive information, which could be identified both linearly and non-linearly. On the other hand, VFAE and the LFR methods were able to retain less information in their latent representation, since only Random Forest was able to achieve higher than random chance accuracy. This further justifies our choice for including the MMD penalty in the lower bound of the VAE. .

\begin{figure}[h]
  \centering
  \begin{subfigure}{.25\textwidth}
        \centering
        \includegraphics[width=1.\textwidth]{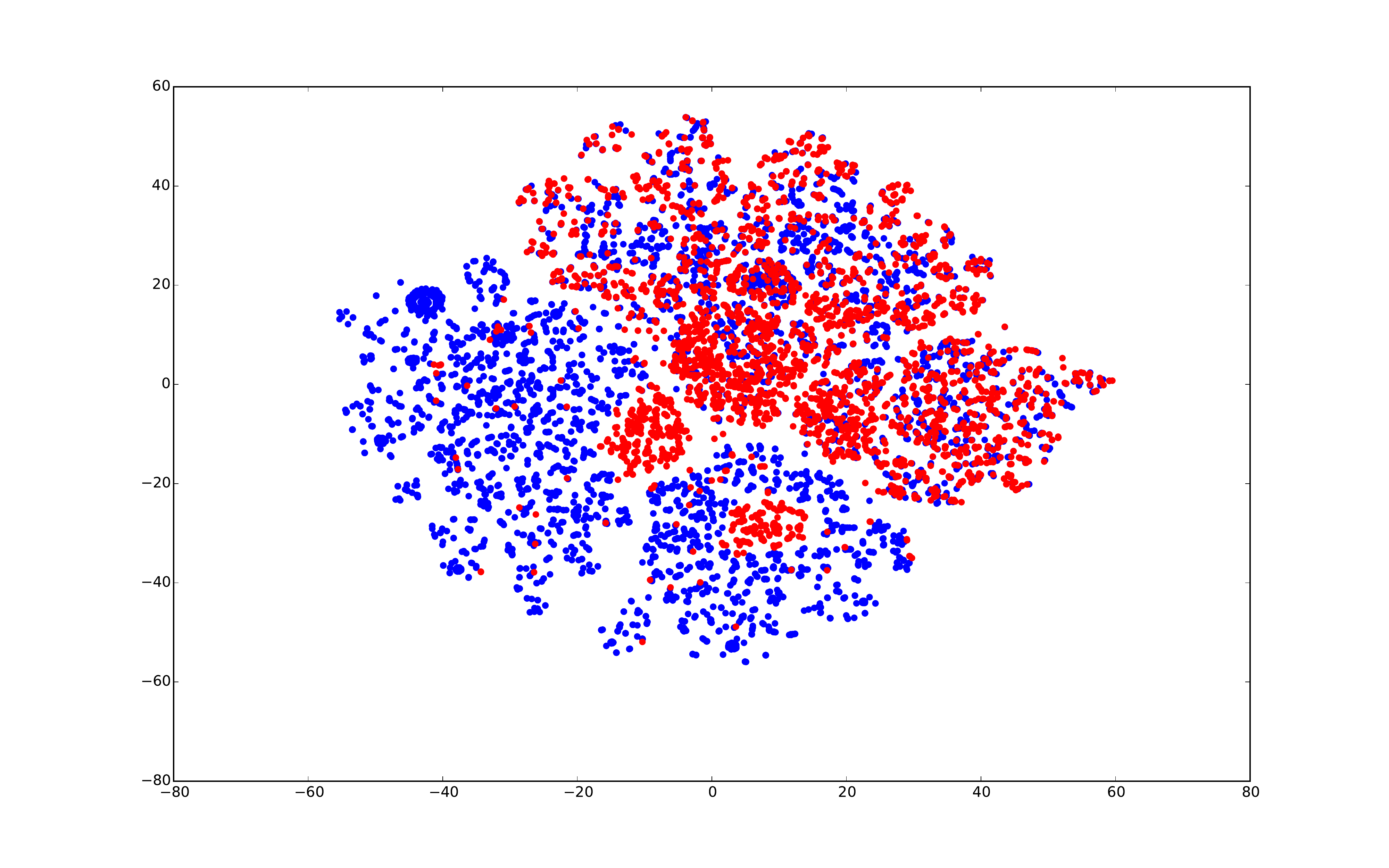}
        \caption{}
        \label{fig:tsne_adult_x}
    \end{subfigure}%
    \begin{subfigure}{.25\textwidth}
        \centering
        \includegraphics[width=1.\textwidth]{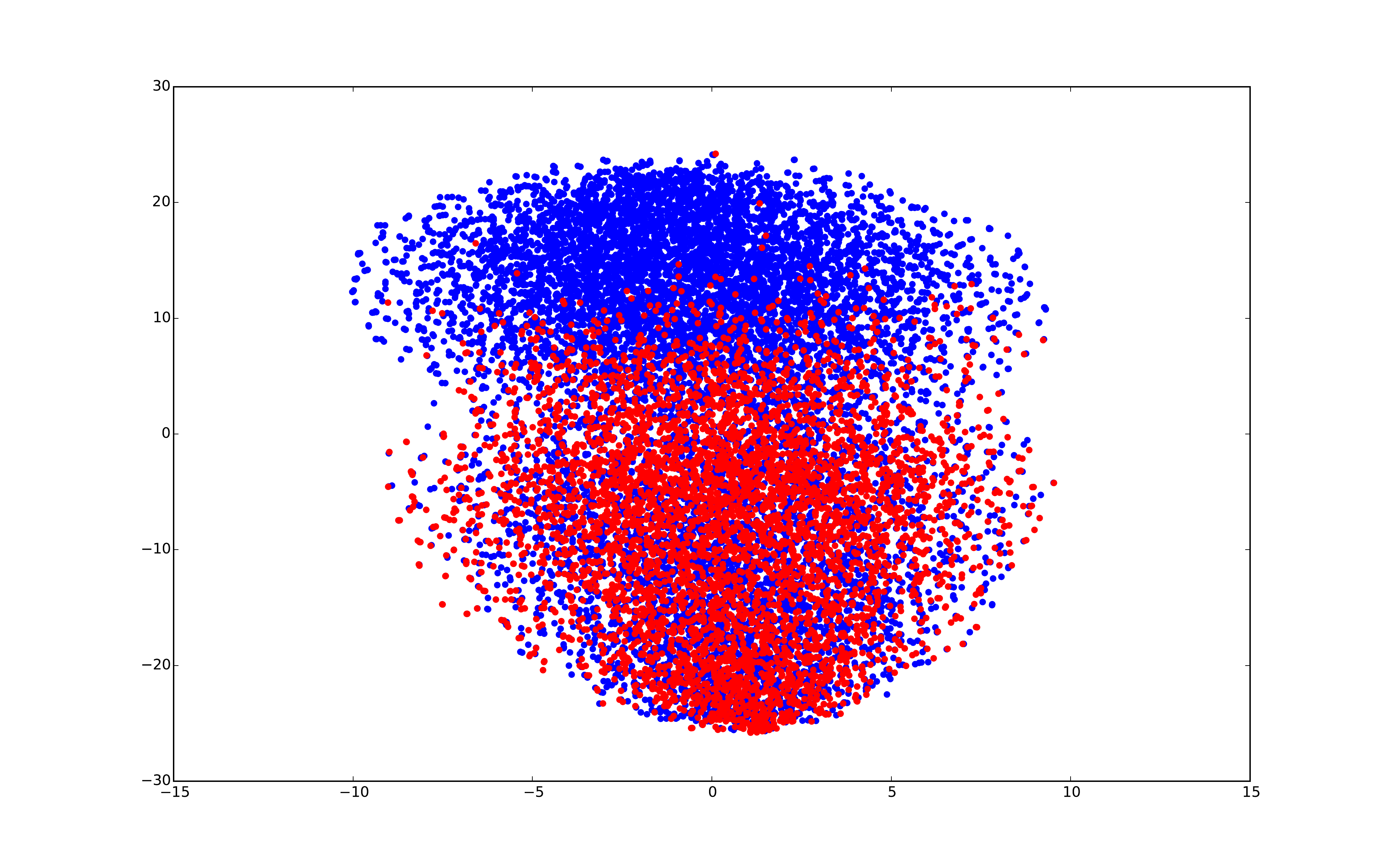}
        \caption{}
        \label{fig:tsne_adult}
    \end{subfigure}%
    \begin{subfigure}{.25\textwidth}
        \centering
        \includegraphics[width=1.\textwidth]{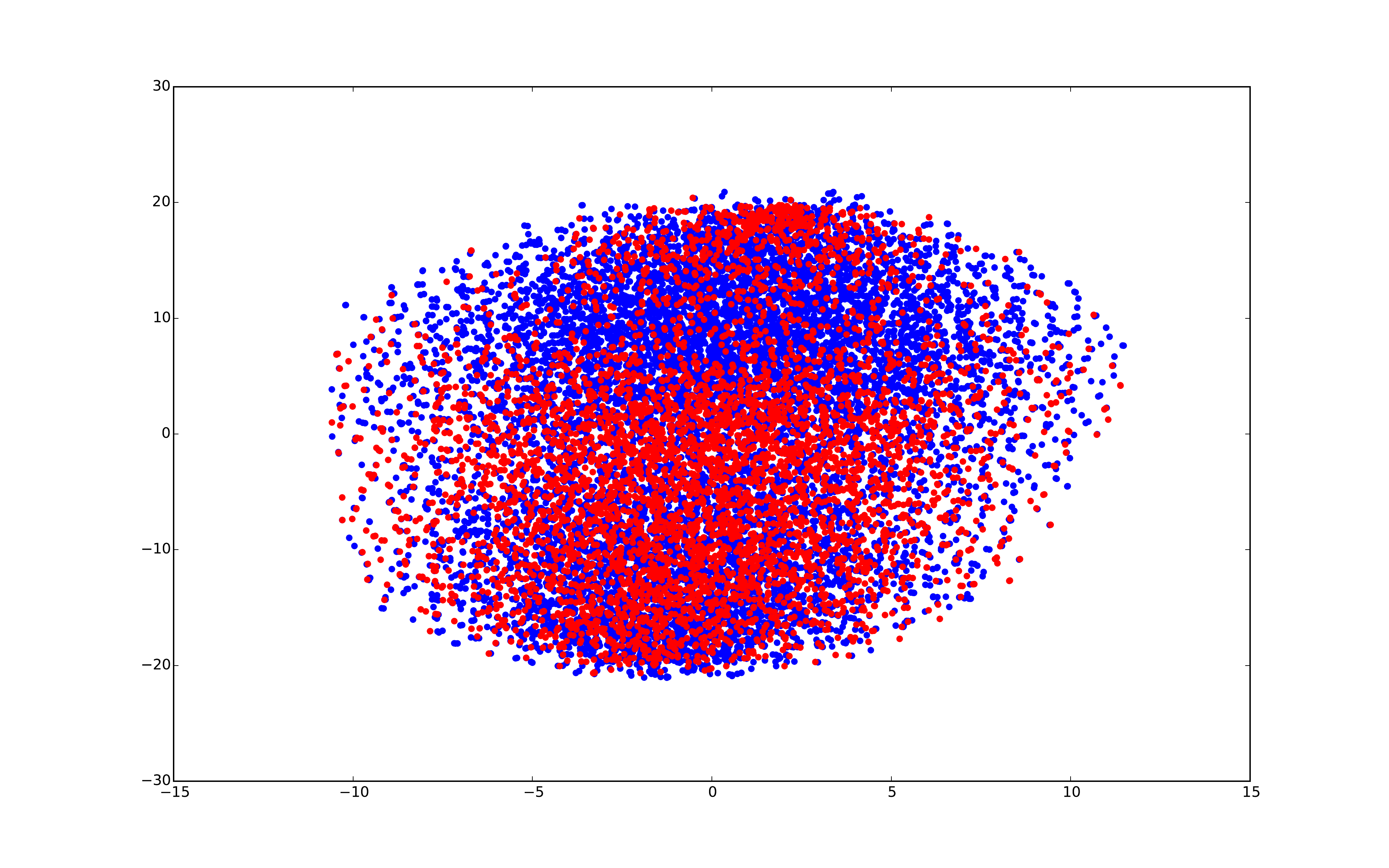}
        \caption{}
        \label{fig:tsne_adult_s}
    \end{subfigure}%
    \begin{subfigure}{.25\textwidth}
        \centering
        \includegraphics[width=1.\textwidth]{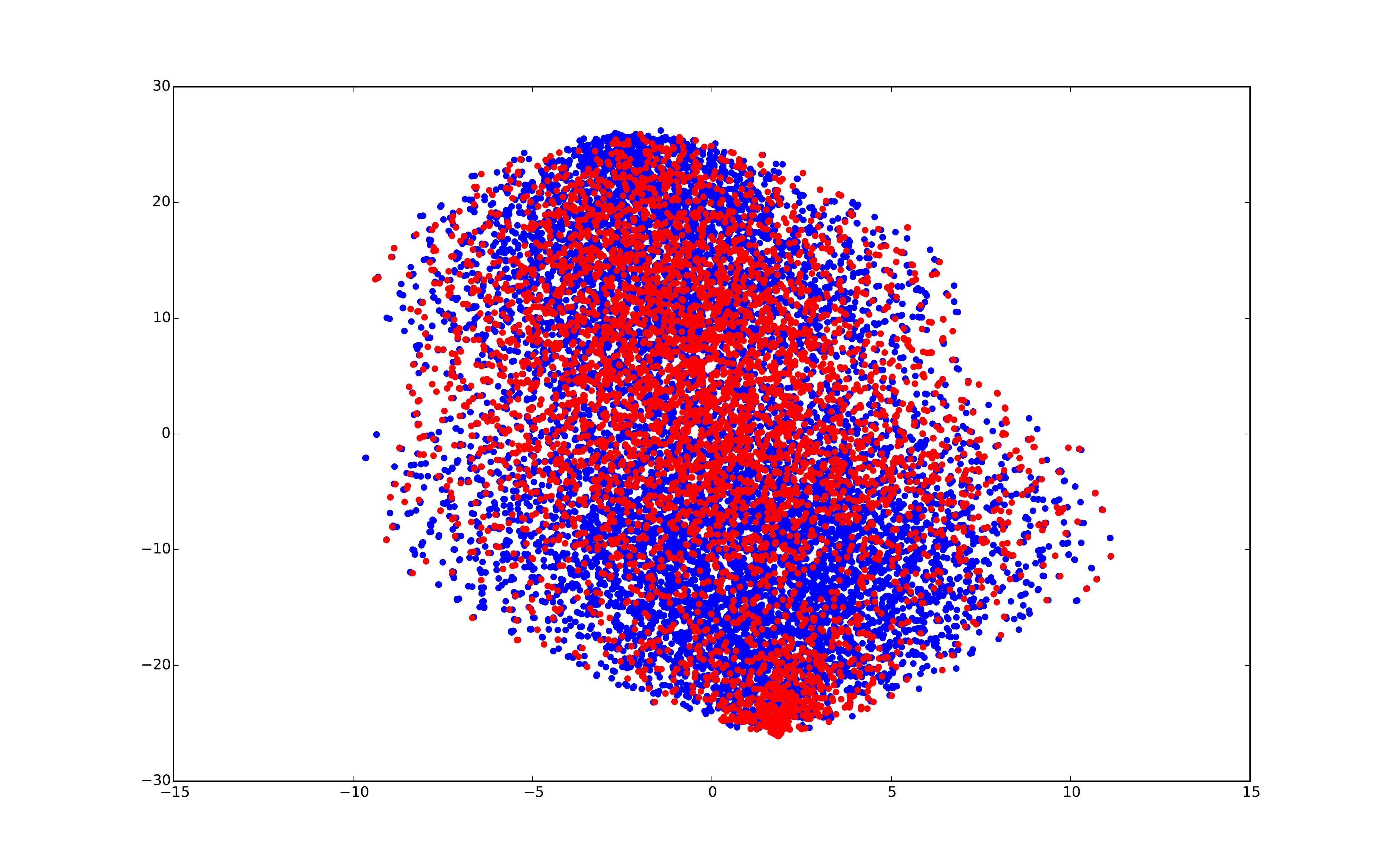}
        \caption{}
        \label{fig:tsne_adult_mmd_s}
    \end{subfigure}%
    \caption{t-SNE~\citep{2013arXiv1301.3342V} visualizations from the Adult dataset on: (a): original $\*x$ , (b):  latent $\*z_1$  without $\*s$ and MMD, (c): latent $\*z_1$ with $\*s$ and without MMD, (d):  latent $\*z_1$ with $\*s$ and MMD. Blue colour corresponds to males whereas red colour corresponds to females.}
    \label{fig:tsne_adult_all}
\end{figure}

In order to further assess the nature of our new representations, we visualized two dimensional Barnes-Hut SNE~\citep{2013arXiv1301.3342V} embeddings of the $\*z_1$ representations, obtained from the model trained on the Adult dataset, in Figure~\ref{fig:tsne_adult_all}. As we can see, the nuisance/sensitive variables $\*s$ can be identified both on the original representation $\*x$ and on a latent representation $\*z_1$ that does not have the MMD penalty and the independence properties between $\*z_1$ and $\*s$ in the prior. By introducing these independence properties as well as the MMD penalty the nuisance variable groups become practically indistinguishable.

\subsubsection{Domain adaptation}
As for the domain adaptation scenario and the Amazon reviews dataset, the results of our VFAE model can be seen in Table~\ref{tab:amazon_results}. Our model was successful in factoring out the domain information, since the accuracy, measured both linearly (LR) and non-linearly (RF), was towards random chance (which for this dataset is 0.5).  We should also mention that, on this dataset at least, completely removing information about the domain does not guarantee a better performance on $\*y$. The same effect was also observed by~\cite{2015arXiv150507818G} and~\cite{chen2012marginalized}. As far as the accuracy on $\*y$ is concerned, we compared against a recent neural network based state of the art method for domain adaptation, Domain Adversarial Neural Network (DANN)~\citep{2015arXiv150507818G}. As we can observe in table~\ref{tab:amazon_results}, our accuracy on the labels $\*y$ is higher on 9 out of the 12 domain adaptation tasks whereas on the remaining 3 it is quite similar to the DANN architecture. 

\begin{table}[ht]
	\caption {Results on the Amazon reviews dataset. The DANN column is taken directly from~\cite{2015arXiv150507818G} (the column that uses the original representation as input).}
	\centering
	\label{tab:amazon_results}
	\begin{center}
		\begin{tabular}{l|l|l|l|l}
			\hline
			\multirow{2}{*}{Source - Target}  & \multicolumn{2}{c|}{S}  & \multicolumn{2}{c}{Y}  \\\cline{2-5}
			& RF & LR & VFAE & DANN \\\hline
			books - dvd & 0.535 & 0.564 & \textbf{0.799} & 0.784 \\ 
			books - electronics & 0.541 & 0.562 & \textbf{0.792} & 0.733 \\
			books - kitchen & 0.537 & 0.583 &  \textbf{0.816} & 0.779 \\
			dvd - books & 0.537 & 0.563 & \textbf{0.755} & 0.723 \\
			dvd - electronics & 0.538 & 0.566 & \textbf{0.786} & 0.754\\
			dvd - kitchen & 0.543 & 0.589 &  \textbf{0.822} & 0.783\\
			electronics - books & 0.562 & 0.590  & \textbf{0.727} & 0.713\\
			electronics - dvd & 0.556 & 0.586 & \textbf{0.765} & 0.738\\
			electronics - kitchen & 0.536 & 0.570 & 0.850 & \textbf{0.854}\\ 
			kitchen - books & 0.560 & 0.593 & \textbf{0.720} & 0.709  \\
			kitchen - dvd & 0.561 & 0.599 & 0.733 & \textbf{0.740} \\
			kitchen - electronics & 0.533 & 0.565  & 0.838 & \textbf{0.843} \\
			\hline
		\end{tabular}
		\end{center}
\end{table}

\subsection{Learning Invariant Representations}
\begin{figure}[ht]
    \centering
    \begin{subfigure}{.49\textwidth}
    \centering
        \includegraphics[width=\textwidth]{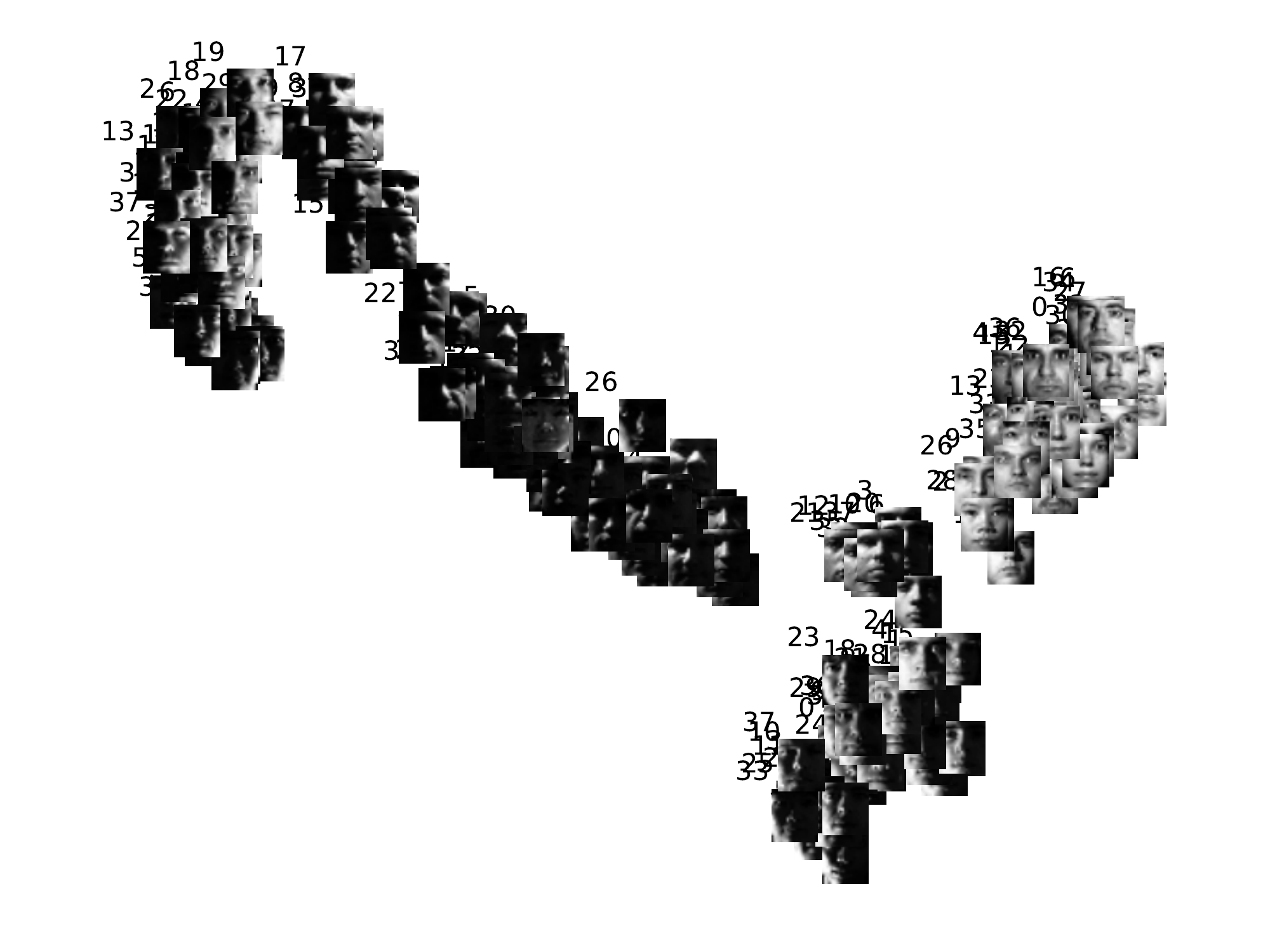}
        \caption{}
        \label{fig:yaleb_x}
  \end{subfigure} %
    \begin{subfigure}{.49\textwidth}
      \centering
        \includegraphics[width=\textwidth]{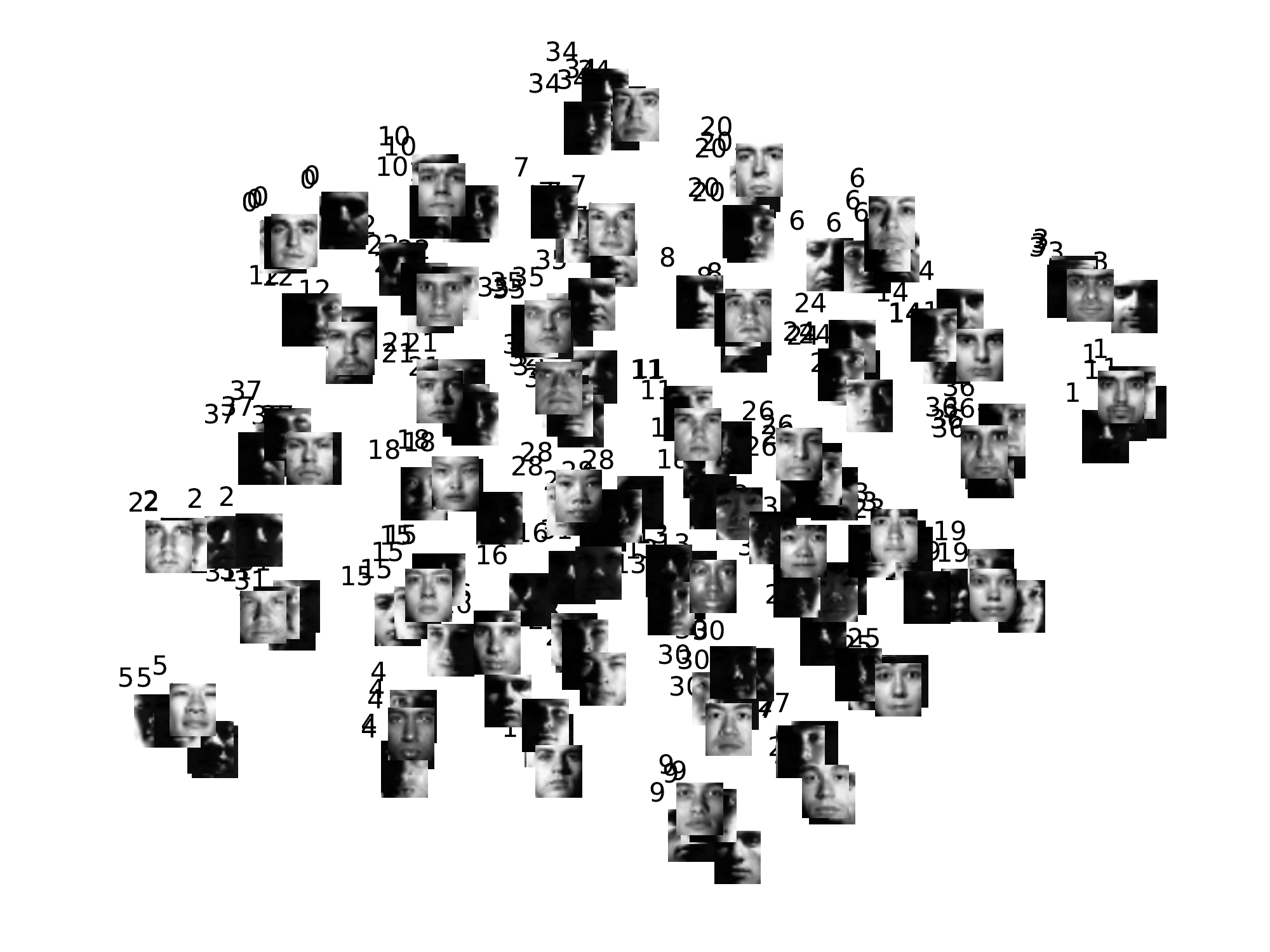}
        \caption{}
        \label{fig:yaleb_z}
     \end{subfigure} %
    \caption{t-SNE~\citep{2013arXiv1301.3342V} visualizations of the Extended Yale B training set. (a): original $\*x$ , (b):  latent $\*z_1$  from VFAE. Each example is plotted with the person ID and the image. Zoom in to see details.}
    \label{fig:yaleb_via}
\end{figure}

Regarding the more general task of learning invariant representations; our results on the Extended Yale B dataset also demonstrate our model's ability to learn such representations. As expected, on the original representation $\*x$ the lighting conditions, $\*s$, are well identifiable with almost perfect accuracy from both RF and LR. This can also be seen in the two dimensional embeddings of the original space $\*x$ in Figure~\ref{fig:yaleb_x}: the images are mostly clustered according to the lighting conditions. As soon as we utilize our VFAE model we simultaneously decrease the accuracy on $\*s$, from 96\% to about 50\%, and increase our accuracy on $\*y$, from 78\% to about 85\%. This effect can also be seen in Figure~\ref{fig:yaleb_z}: the images are now mostly clustered according to the person ID (the label $\*y$). It is clear that in this scenario the information about  $\*s$ is purely ``nuisance'' with respect to the labels $\*y$. Therefore, by using our VFAE model we are able to obtain improved generalization and classification performance by effectively removing $\*s$ from our representations.
   
\begin{table}[ht]
	\caption {Results on the Extended Yale B dataset. We also included the best result from~\cite{li2014learning} under the NN + MMD row.}
	\centering
	\label{tab:yaleb_results}
	\begin{center}
		\begin{tabular}{l|l|l|l}
			\hline
			\multirow{2}{*}{Method}  & \multicolumn{2}{c|}{S}  &  \multirow{2}{*}{Y} \\\cline{2-3}
			& RF & LR \\\hline
			Original $\*x$ & 0.952 & 0.961 & 0.78\\ 
			NN + MMD & - & - & 0.82\\
			VFAE & 0.435 & 0.565 & \textbf{0.846}\\
			\hline
		\end{tabular}
		\end{center}
\end{table}

\section{Related Work}
Most related to our ``fair'' representations view is the work from~\cite{zemel2013learning}. They proposed a neural network based semi-supervised clustering model for learning fair representations. The idea is to learn a localised representation that maps each datapoint to a cluster in such a way that each cluster gets assigned roughly equal proportions of data from each group in $s$. Although their approach was successfully applied on several datasets, the restriction to clustering means that it cannot leverage the representational power of a distributed representation. Furthermore, this penalty does not account for higher order moments in the latent distribution. For example, if $p(z_k=1 | x_i, s=0)$ always returns $1$ or $0$, while $p(z_k=1 | x_i, s=1)$ returns values between values $0$ and $1$, then the penalty could still be satisfied, but information could still leak through. We addressed both of these issues in this paper.

Domain adaptation can also be cast as learning representations that are ``invariant'' with respect to a discrete variable $\*s$, the domain. Most similar to our work are neural network approaches which try to match the feature distributions between the domains. This was performed in an unsupervised way with mSDA~\citep{chen2012marginalized} by training denoising autoencoders jointly on all domains, thus implicitly obtaining a representation general enough to explain both the domain and the data. This is in contrast to our approach where we instead try to learn representations that explicitly remove domain information during the learning process. For the latter we find more similarities with ``domain-regularized'' supervised approaches that simultaneously try to predict the label for a data point and remove domain specific information. This is done with either MMD~\citep{long2015learning, DBLP:journals/corr/TzengHZSD14} or adversarial~\citep{2015arXiv150507818G} penalties at the hidden layers of the network. In our model however the main ``domain-regularizer'' stems from the independence properties of the prior over the domain and latent representations. We also employ MMD on our model but from a different perspective since we consider a slightly more difficult case where the domain $\*s$ and label $\*y$ are correlated; we need to ensure that we remain as ``invariant'' as possible since $q_\phi(\*y|\*z_1)$ might `leak' information about $\*s$. 

\section{Conclusion}
We introduce the Variational Fair Autoencoder (VFAE), an extension of the semi-supervised variational autoencoder in order to learn representations that are explicitly invariant with respect to some known aspect of a dataset while retaining as much remaining information as possible. We further use a Maximum Mean Discrepancy regularizer in order to further promote invariance in the posterior distribution over latent variables. We apply this model to tasks involving developing fair classifiers that are invariant to sensitive demographic information and show that it produces a better tradeoff with respect to accuracy and invariance. As a second application, we consider the task of domain adaptation, where the goal is to improve classification by training a classifier that is invariant to the domain. We find that our model is competitive with recently proposed adversarial approaches. Finally, we also consider the more general task of learning invariant representations. We can observe that our model provides a clear improvement against a neural network that incorporates a Maximum Mean Discrepancy penalty.


\bibliography{bibliography}

\begin{thebibliography}{18}
\providecommand{\natexlab}[1]{#1}
\providecommand{\url}[1]{\texttt{#1}}
\expandafter\ifx\csname urlstyle\endcsname\relax
  \providecommand{\doi}[1]{doi: #1}\else
  \providecommand{\doi}{doi: \begingroup \urlstyle{rm}\Url}\fi

\bibitem[Ben-David et~al.(2007)Ben-David, Blitzer, Crammer, Pereira,
  et~al.]{ben2007analysis}
Ben-David, Shai, Blitzer, John, Crammer, Koby, Pereira, Fernando, et~al.
\newblock Analysis of representations for domain adaptation.
\newblock \emph{Advances in neural information processing systems},
  19:\penalty0 137, 2007.

\bibitem[Ben-David et~al.(2010)Ben-David, Blitzer, Crammer, Kulesza, Pereira,
  and Vaughan]{ben2010theory}
Ben-David, Shai, Blitzer, John, Crammer, Koby, Kulesza, Alex, Pereira,
  Fernando, and Vaughan, Jennifer~Wortman.
\newblock A theory of learning from different domains.
\newblock \emph{Machine learning}, 79\penalty0 (1-2):\penalty0 151--175, 2010.

\bibitem[Chen et~al.(2012)Chen, Xu, Weinberger, and Sha]{chen2012marginalized}
Chen, Minmin, Xu, Zhixiang, Weinberger, Kilian, and Sha, Fei.
\newblock Marginalized denoising autoencoders for domain adaptation.
\newblock \emph{International Conference on Machine Learning (ICML)}, 2012.

\bibitem[Frank \& Asuncion(2010)Frank and Asuncion]{UCI}
Frank, A. and Asuncion, A.
\newblock {UCI} machine learning repository, 2010.
\newblock URL \url{http://archive.ics.uci.edu/ml}.

\bibitem[{Ganin} et~al.(2015){Ganin}, {Ustinova}, {Ajakan}, {Germain},
  {Larochelle}, {Laviolette}, {Marchand}, and {Lempitsky}]{2015arXiv150507818G}
{Ganin}, Y., {Ustinova}, E., {Ajakan}, H., {Germain}, P., {Larochelle}, H.,
  {Laviolette}, F., {Marchand}, M., and {Lempitsky}, V.
\newblock {Domain-Adversarial Training of Neural Networks}.
\newblock \emph{ArXiv e-prints}, May 2015.

\bibitem[Gretton et~al.(2006)Gretton, Borgwardt, Rasch, Sch{\"o}lkopf, and
  Smola]{gretton2006kernel}
Gretton, Arthur, Borgwardt, Karsten~M, Rasch, Malte, Sch{\"o}lkopf, Bernhard,
  and Smola, Alex~J.
\newblock A kernel method for the two-sample-problem.
\newblock In \emph{Advances in neural information processing systems}, pp.\
  513--520, 2006.

\bibitem[Kifer et~al.(2004)Kifer, Ben-David, and Gehrke]{kifer2004detecting}
Kifer, Daniel, Ben-David, Shai, and Gehrke, Johannes.
\newblock Detecting change in data streams.
\newblock In \emph{Proceedings of the Thirtieth international conference on
  Very large data bases-Volume 30}, pp.\  180--191. VLDB Endowment, 2004.

\bibitem[Kingma \& Ba(2015)Kingma and Ba]{DBLP:journals/corr/KingmaB14}
Kingma, Diederik~P and Ba, Jimmy.
\newblock Adam: {A} method for stochastic optimization.
\newblock \emph{International Conference on Learning Representations (ICLR)},
  2015.

\bibitem[Kingma \& Welling(2014)Kingma and Welling]{kingma2013auto}
Kingma, Diederik~P and Welling, Max.
\newblock Auto-encoding variational bayes.
\newblock \emph{International Conference on Learning Representations (ICLR)},
  2014.

\bibitem[Kingma et~al.(2014)Kingma, Mohamed, Rezende, and
  Welling]{kingma2014semi}
Kingma, Diederik~P, Mohamed, Shakir, Rezende, Danilo~Jimenez, and Welling, Max.
\newblock Semi-supervised learning with deep generative models.
\newblock In \emph{Advances in Neural Information Processing Systems}, pp.\
  3581--3589, 2014.

\bibitem[Li et~al.(2014)Li, Swersky, and Zemel]{li2014learning}
Li, Yujia, Swersky, Kevin, and Zemel, Richard.
\newblock Learning unbiased features.
\newblock \emph{arXiv preprint arXiv:1412.5244}, 2014.

\bibitem[Long \& Wang(2015)Long and Wang]{long2015learning}
Long, Mingsheng and Wang, Jianmin.
\newblock Learning transferable features with deep adaptation networks.
\newblock \emph{arXiv preprint arXiv:1502.02791}, 2015.

\bibitem[Rahimi \& Recht(2009)Rahimi and Recht]{rahimi2009weighted}
Rahimi, Ali and Recht, Benjamin.
\newblock Weighted sums of random kitchen sinks: Replacing minimization with
  randomization in learning.
\newblock In \emph{Advances in neural information processing systems}, pp.\
  1313--1320, 2009.

\bibitem[Rezende et~al.(2014)Rezende, Mohamed, and
  Wierstra]{rezende2014stochastic}
Rezende, Danilo~Jimenez, Mohamed, Shakir, and Wierstra, Daan.
\newblock Stochastic backpropagation and approximate inference in deep
  generative models.
\newblock \emph{International Conference on Machine Learning (ICML)}, 2014.

\bibitem[Tzeng et~al.(2014)Tzeng, Hoffman, Zhang, Saenko, and
  Darrell]{DBLP:journals/corr/TzengHZSD14}
Tzeng, Eric, Hoffman, Judy, Zhang, Ning, Saenko, Kate, and Darrell, Trevor.
\newblock Deep domain confusion: Maximizing for domain invariance.
\newblock \emph{CoRR}, abs/1412.3474, 2014.
\newblock URL \url{http://arxiv.org/abs/1412.3474}.

\bibitem[{van der Maaten}(2013)]{2013arXiv1301.3342V}
{van der Maaten}, L.
\newblock {Barnes-Hut-SNE}.
\newblock \emph{ArXiv e-prints}, January 2013.

\bibitem[Zemel et~al.(2013)Zemel, Wu, Swersky, Pitassi, and
  Dwork]{zemel2013learning}
Zemel, Rich, Wu, Yu, Swersky, Kevin, Pitassi, Toni, and Dwork, Cynthia.
\newblock Learning fair representations.
\newblock In \emph{Proceedings of the 30th International Conference on Machine
  Learning (ICML-13)}, pp.\  325--333, 2013.

\bibitem[Zhao \& Meng(2015)Zhao and Meng]{zhao2014fastmmd}
Zhao, Ji and Meng, Deyu.
\newblock Fastmmd: Ensemble of circular discrepancy for efficient two-sample
  test.
\newblock \emph{Neural computation}, 2015.

\end{thebibliography}
\bibliographystyle{iclr2016_conference}

\newpage
\appendix

\section{Discrimination metrics}
The Discrimination metric~\citep{zemel2013learning} and the Discrimination metric that takes into account the probabilities of the correct class are mathematically formalized as:
 
\begin{align*}
 \text{Discrimination} & = \bigg|\frac{\sum_{n=1}^{N}\mathbb{I}[y_n^{s=0}]}{N_{s = 0}} - \frac{\sum_{n=1}^{N}\mathbb{I}[y_n^{s=1}]}{N_{s=1}}\bigg| \\  
  \text{Discrimination prob.} & = \bigg|\frac{\sum_{n=1}^{N}p(y_n^{s=0})}{N_{s = 0}} - \frac{\sum_{n=1}^{N}p(y_n^{s=1})}{N_{s=1}}\bigg|
\end{align*}
where $\mathbb{I}[y_n^{s=0}] = 1$ for the predictions $y_n$ that were done on the datapoints with nuisance variable $s = 0$, $N_{s=0}$ denotes the total amount of datapoints that had nuisance variable $s = 0$ and $p(y_n^{s=0})$ denotes the probability of the prediction $y_n$ for the datapoints with $s=0$. For the predictions and their respective probabilities we used a Logistic Regression classifier.

\section{Proxy A-Distance (PAD) for Amazon Reviews dataset}
\begin{figure}[ht]
    \centering
    \begin{subfigure}{.49\textwidth}
    \centering
        \includegraphics[width=1.\textwidth]{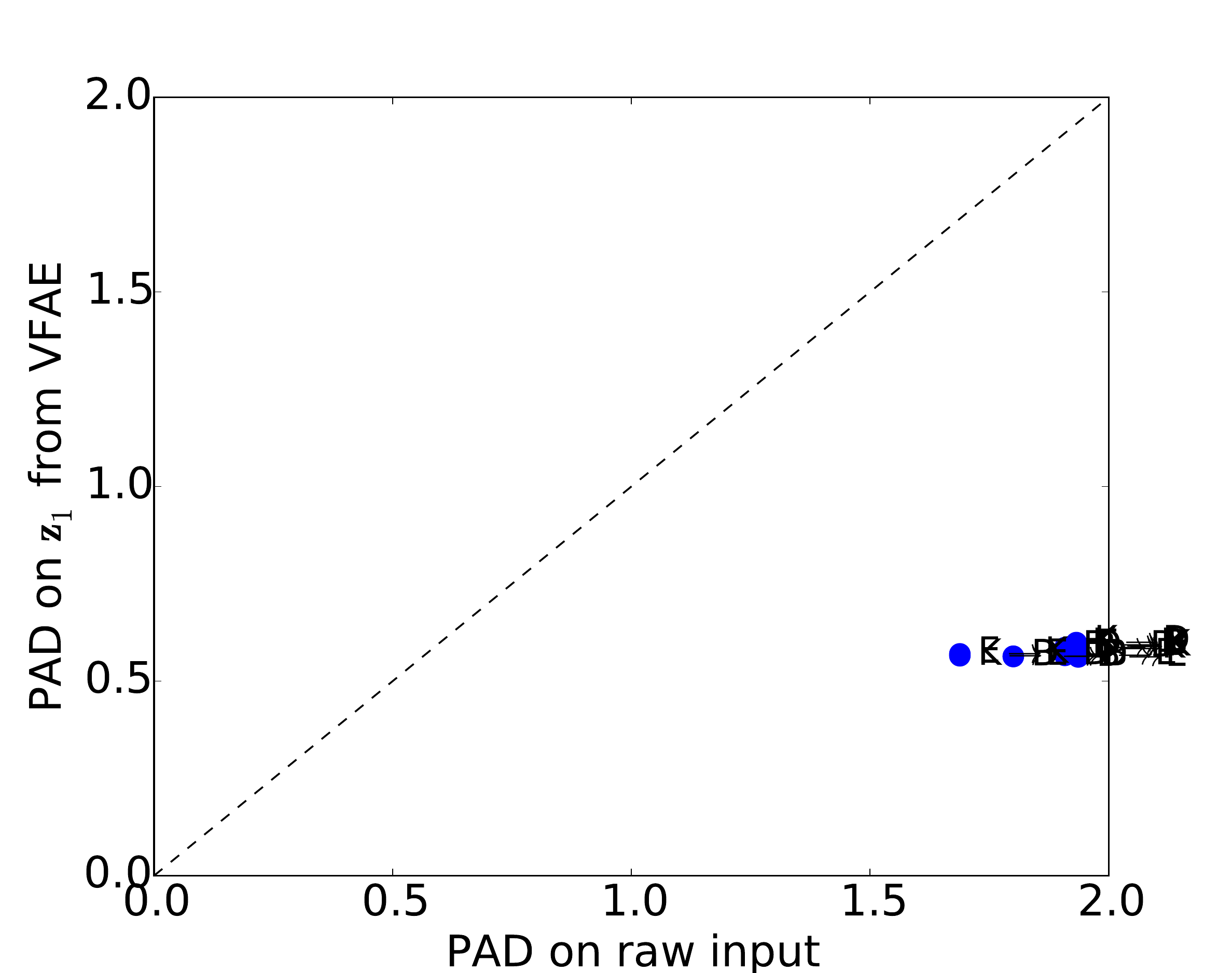}
  \end{subfigure} %
    \begin{subfigure}{.49\textwidth}
      \centering
        \includegraphics[width=.9\textwidth]{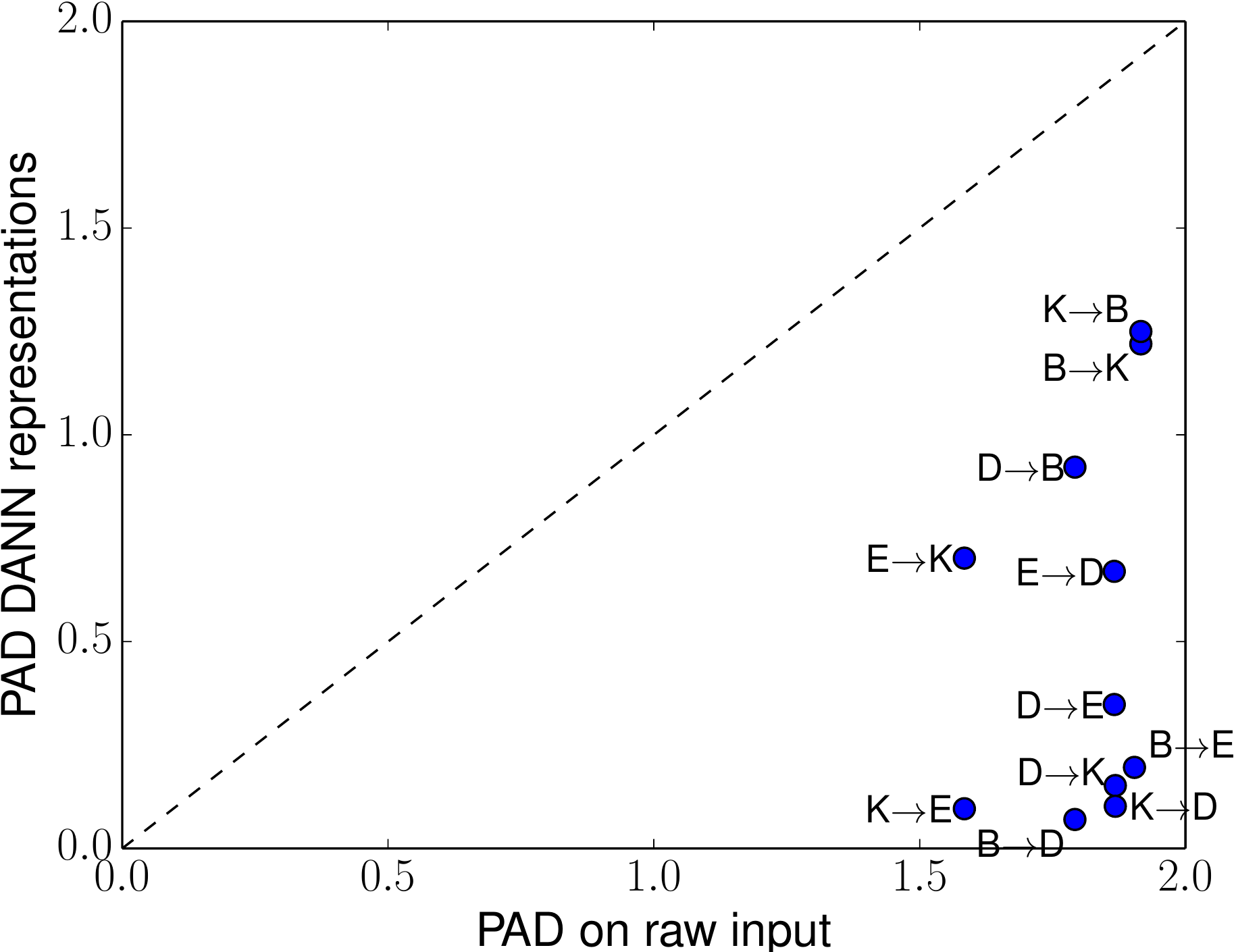}
     \end{subfigure} %
    \caption{Proxy A-distances (PAD) for the Amazon reviews dataset: left from our VFAE model, right from the DANN model (taken from~\cite{2015arXiv150507818G})}
    \label{fig:pad_reviews}
\end{figure}

Similarly to~\cite{2015arXiv150507818G}, we also calculated the Proxy A-distance (PAD)~\citep{ben2007analysis,ben2010theory} scores for the raw data $\*x$ and for the $\*z_1$ representations of VFAE. Briefly, Proxy A-distance is an approximation to the $\mathcal{H}$-divergence measure of domain distinguishability proposed in~\cite{kifer2004detecting} and~\cite{ben2007analysis,ben2010theory}. To compute it we first need to train a learning algorithm on the task of discriminating examples from the source and target domain. Afterwards we can use the test error $\epsilon$ of that algorithm in the following formula:
\begin{align*}
  \text{PAD}(\epsilon)  = 2(1 - 2\epsilon) 
\end{align*}
It is clear that low PAD scores correspond to low discrimination of the source and target domain examples from the classifier. To obtain $\epsilon$ for our model we used Logistic Regression. The resulting plot can be seen in Figure~\ref{fig:pad_reviews}, where we have also added the plot from DANN~\citep{2015arXiv150507818G}, where they used a linear Support Vector Machine for the classifier, as a reference. It can be seen that our VFAE model can factor out the information about $\*s$ better, since the PAD scores on our new representation are, overall, lower than the ones obtained from the DANN architecture.

\end{document}